\documentclass[journal]{IEEEtran}
\usepackage{courier}
\usepackage{amsmath}
\usepackage{amssymb}
\usepackage{amsfonts}
\usepackage{bbding}
\usepackage{balance}
\usepackage{indentfirst}
\usepackage{xcolor}
\usepackage{multirow}
\usepackage{float}
\usepackage{graphicx}
\usepackage{graphics}
\usepackage{subfigure}
\usepackage{float}
\usepackage{bigstrut}
\usepackage{makecell}
\usepackage{booktabs}
\usepackage{bibentry}
\usepackage{epstopdf}
\usepackage{wasysym}
\usepackage{tabularx}
\usepackage{comment}
\usepackage{algorithm}
\usepackage{algpseudocode}
\usepackage{pifont}
\usepackage{url}
\usepackage{fixltx2e}
\usepackage{placeins}

\newcommand{\RED}[1]{\textcolor{black}{#1}}

 \makeatletter
    \newcommand{\thickhline}{%
        \noalign {\ifnum 0=`}\fi \hrule height 1pt
        \futurelet \reserved@a \@xhline
    }
    \newcolumntype{"}{@{\vrule width 1pt}}
    \makeatother
\makeatletter
\let\ftype@table\ftype@figure
\makeatother

\let\oldFootnote\footnote
\newcommand\nextToken\relax

\renewcommand\footnote[1]{%
    \oldFootnote{#1}\futurelet\nextToken\isFootnote}

\newcommand\isFootnote{%
    \ifx\footnote\nextToken\textsuperscript{,}\fi}

\ifCLASSINFOpdf
\else
\fi

\begin{document}

\title{Efficient Token-Guided Image-Text Retrieval with Consistent Multimodal Contrastive Training}
\author{Chong~Liu$^{*}$, Yuqi~Zhang$^{*}$, Hongsong~Wang$^{*\dagger}$, Weihua~Chen$^{\dagger}$, Fan Wang, Yan Huang, Yi-Dong Shen,
        and~Liang~Wang, Fellow, IEEE 
\IEEEcompsocitemizethanks{
\IEEEcompsocthanksitem Corresponding author: Hongsong~Wang, Weihua Chen. *~indicates equal contributions. 
\IEEEcompsocthanksitem C.~Liu and Y.~Shen are with State Key Laboratory of Computer Science, Institute of Software, Chinese Academy of Sciences, Beijing 100190, China (e-mail: liuchong@ios.ac.cn; ydshen@ios.ac.cn).
\IEEEcompsocthanksitem Y.~Zhang, W.~Chen and F.~Wang are from Alibaba Group, Beijing 100102, China (e-mail: gongyou.zyq@alibaba-inc.com; kugang.cwh@alibaba-inc.com; fan.w@alibaba-inc.com).
\IEEEcompsocthanksitem H.~Wang is with Department of Computer Science and Engineering, Southeast University, Nanjing 210096, China (e-mail: hongsongwang@seu.edu.cn).
\IEEEcompsocthanksitem Y.~Huang and L.~Wang are with the Center for Research on Intelligent Perception and Computing (CRIPAC), National Laboratory of Pattern Recognition (NLPR), Institute of Automation, Chinese Academy of Sciences (CASIA), Beijing 100190, China (e-mail: yhuang@nlpr.ia.ac.cn; wangliang@nlpr.ia.ac.cn).
 }
}

\markboth{JOURNAL OF LATEX CLASS FILES,~Vol.~xx, No.~xx, xx~2017}%
{Shell \MakeLowercase{\textit{et al.}}: Bare Demo of IEEEtran.cls for IEEE Journals}

\maketitle

\begin{abstract}
Image-text retrieval is a central problem for understanding the semantic relationship between vision and language, and serves as the basis for various visual and language tasks. Most previous works either simply learn coarse-grained representations of the overall image and text, or elaborately establish the correspondence between image regions or pixels and text words. However, the close relations between coarse- and fine-grained representations for each modality are important for image-text retrieval but almost neglected. As a result, such previous works inevitably suffer from low retrieval accuracy or heavy computational cost. In this work, we address image-text retrieval from a novel perspective by combining coarse- and fine-grained representation learning into a unified framework. This framework is consistent with human cognition, as humans simultaneously pay attention to the entire sample and regional elements to understand the semantic content. 
To this end, a Token-Guided Dual Transformer (TGDT) architecture which consists of two homogeneous branches for image and text modalities, respectively, is proposed for image-text retrieval. The TGDT incorporates both coarse- and fine-grained retrievals into a unified framework and beneficially leverages the advantages of both retrieval approaches. A novel training objective called Consistent Multimodal Contrastive (CMC) loss is proposed accordingly to ensure the intra- and inter-modal semantic consistencies between images and texts in the common embedding space. Equipped with a two-stage inference method based on the mixed global and local cross-modal similarity, the proposed method achieves state-of-the-art retrieval performances with extremely low inference time when compared with representative recent approaches. 
\RED{Code is publicly available: \textcolor{red}{\url{github.com/LCFractal/TGDT}}.}
\end{abstract}

\maketitle
\IEEEdisplaynontitleabstractindextext
\IEEEpeerreviewmaketitle

\section{Introduction} \label{sec:intro}
\IEEEPARstart{I}{mage-text} retrieval aims to match images and texts based on content-based semantic similarities between them. \RED{It is highly relevant to various computer vision tasks and machine learning approaches}, such as image captioning~\cite{2018Bottom}, text-to-image synthesis~\cite{xu2018attngan}, activity understanding\RED{~\cite{zhang2022tn,Wang2018BeyondJL}}, \RED{multimodal machine translation~\cite{li2022video}, scene graph generation~\cite{chang2021comprehensive} and zero-shot learning~\cite{wang2019survey,yan2021zeronas}}. Recently, the task has popularly and continuously attracted attention from both the academic literature and industry. 
However, the semantic gap between image content and linguistic description has always been a major obstacle to the developing of practical retrieval systems.

Image-text retrieval consists of two closely related tasks: text-to-image retrieval and image-to-text retrieval. The first aims to select the image that best matches the given text from the image candidate set, and the latter attempts to find the sentence in the text candidate set that best describes the image. Numerous works have focused on this field and significant progress has been made. We roughly divide previous works into three categories: coarse-grained retrieval, fine-grained retrieval, and vision-language pre-training.

\begin{figure}[t]
	\begin{center}
		\includegraphics[width=0.85\linewidth]{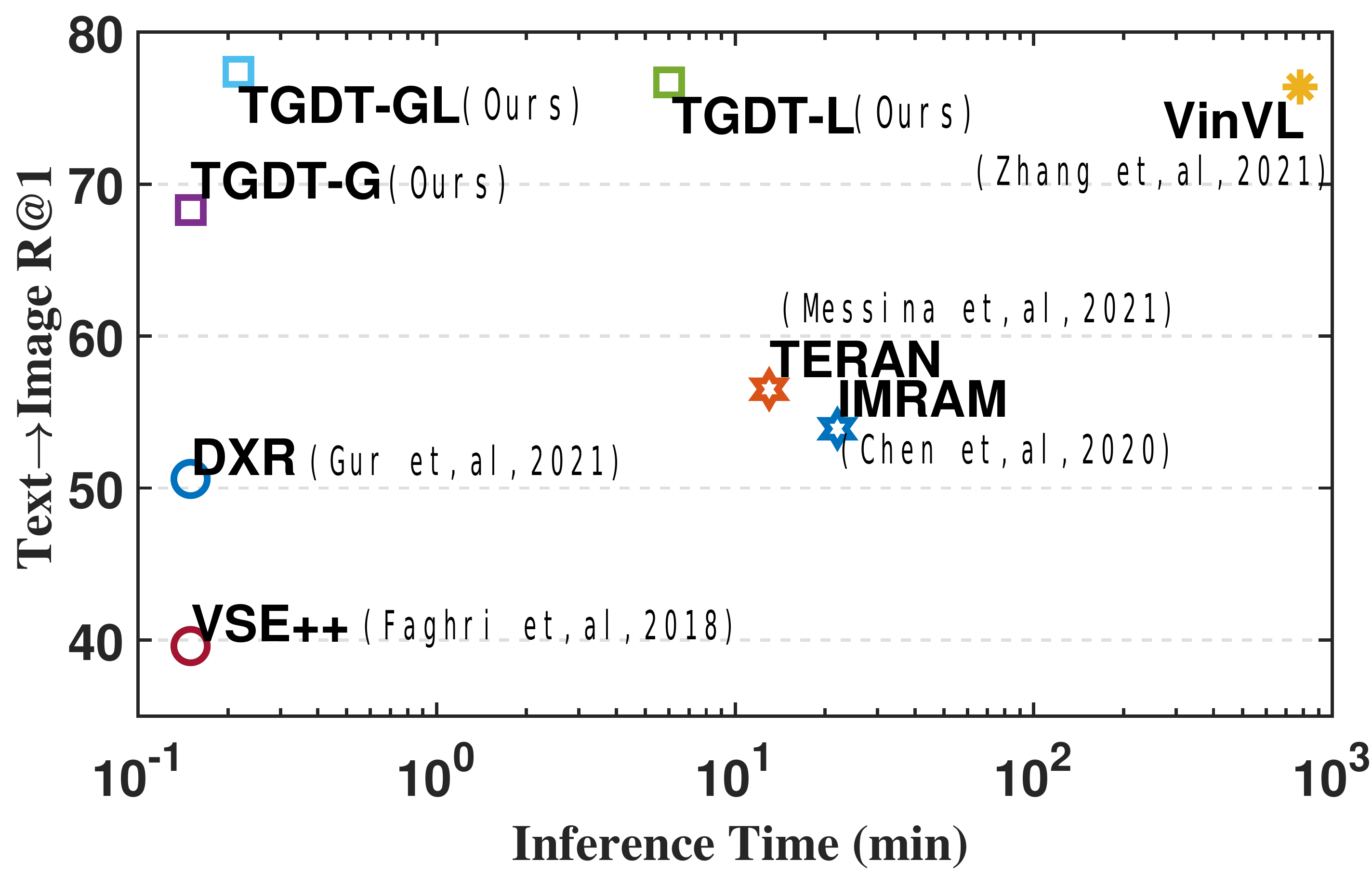}
	\end{center}
  \vspace{-1 em}
	\caption{Comparison of inference time and performance on the Flickr30K dataset. For convenience, we only show the R@1 values of text-to-image retrieval. The inference time represents the total time it takes for image-text retrieval of all samples in the test subset. DXR~\cite{DXR} and VSE++~\cite{VSE} are course-grained retrieval methods. TERAN~\cite{TERAN} and IMRAM~\cite{IMRAM} are fine-grained retrieval methods. VinVL~\cite{VinVL} is vision-language pre-training method.
	} 
	\label{fig:speed_acc}
	 \vspace{-1 em}
\end{figure}
\begin{figure*}[t]
	\begin{center}
		\includegraphics[width=0.9\linewidth]{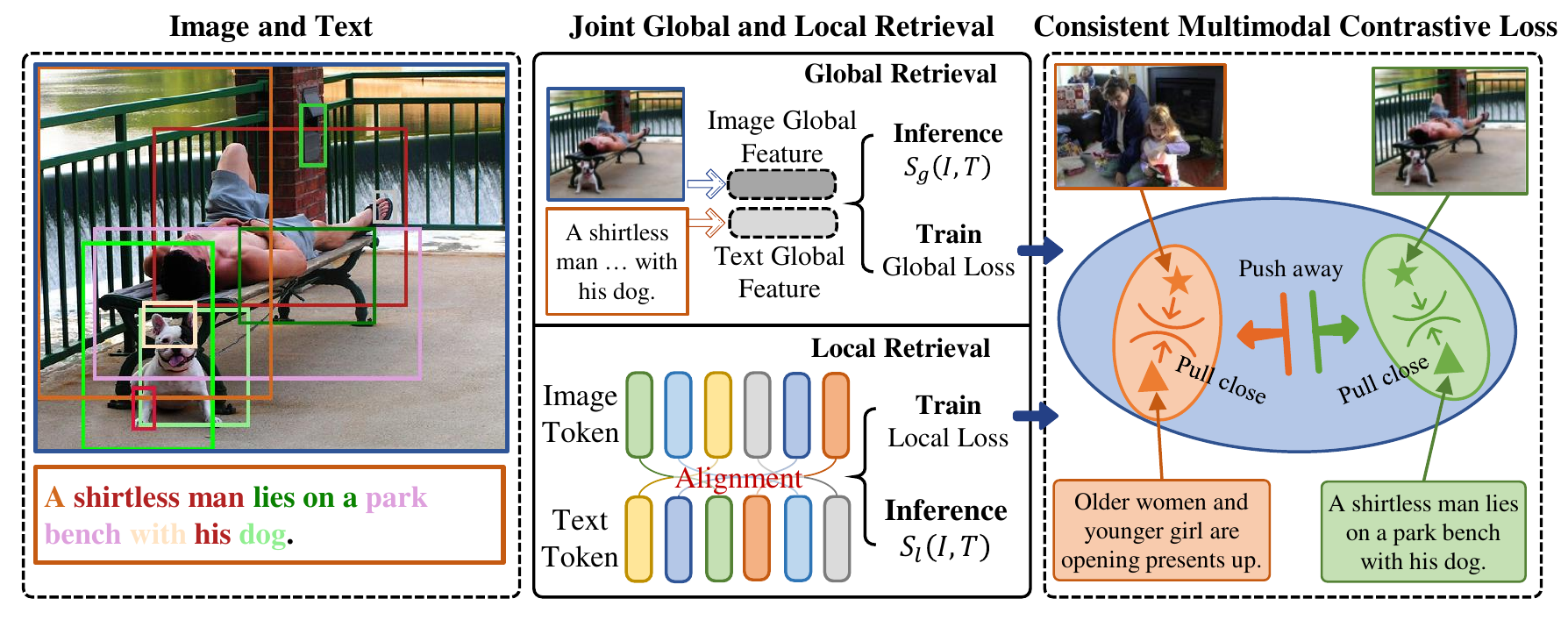}
	\end{center}
   \vspace{-1 em}
	\caption{Outline of the proposed framework for image-text retrieval. The global retrieval module captures holistic representations of the input image and text, while the local retrieval module aligns between image tokens and text tokens. During training, the proposed consistent multimodal contrastive loss jointly optimizes the two modules as well as network structures for both image and text modalities.} 
	\vspace{-0.5 em}
	\label{fig:introduction}
\end{figure*}
The coarse-grained retrieval simply calculates the global similarity between the image and text based on their global representations. Early works~\cite{wang2016learning,DPC,VSE,dong2019dual,wang2018learning,wu2017sampling} obtains this similarity between samples of two modalities by directly mapping the complete image and text into a common embedding space. DPC~\cite{DPC}, VSE++~\cite{VSE} and other works use a deep network with two branches to map the image and text into the embedding space, respectively. However, these works only roughly capture the global correspondence between modalities, and lack the fine-grained interaction between vision and language. Although most of these approaches possess fast inference and good potential scalability for large-scale applications, they could rarely reduce the semantic gap between complex images and texts.

The fine-grained retrieval~\cite{karpathy2017deep,niu2017hierarchical,SCAN,CAMP,VSRN,IMRAM,TERAN,SGRAF,wray2019fine} becomes popular in recent years by exploring the local correspondence between vision and language. The attention mechanisms are usually introduced to align elements of the two modalities. Representative approaches such as SCAN~\cite{SCAN}, IMRAM~\cite{IMRAM}, \emph{etc.}, use the cross-attention to dynamically align each element with all other
elements from the other modality. Since these methods establish the correspondence between parts of each sample between different modalities, they regularly achieve superior performance over the coarse-grained retrieval methods. However, cross attention requires excessive cross computations between an image and a text. The computational inefficiency of these approaches affects the scalability and flexibility for realistic applications.

More recently, Vision-Language Pre-training (VLP) methods based on Transformer gain increasing concerns in the field of multimodal learning. The VLP methods attempt to build a unified model for a wide range of vision-language tasks, typical approaches include Unicoder-VL~\cite{UnicoderVL}, Oscar~\cite{Oscar} and VinVL~\cite{VinVL}. These methods exhibit good generalizability and perform well for the downstream task of image-text retrieval. However, training these models requires vast amounts of external labeled data from different vision-language tasks. In addition, the cross-modal understanding module of the VLP methods requires extensive calculation while modeling modality interactions.

Despite being broadly studied, image-text retrieval remains a challenging problem due to the complexity of each modality and the significant semantic gap between different modalities. Besides, effective cross-modal retrieval approaches with high efficiency are imperative for deployment in realistic industrial scenarios. In brief, most previous efforts either focus on coarse-grained image and text representations or fine-grained local cross-modal correspondences between regions or tokens. A unified model that fully takes advantage of both coarse- and fine-grained approaches has not yet been explored. This learning paradigm is also in line with human perception, as humans simultaneously consider both global and local semantic information retrieving between images and texts.

Motivated by the above analyses, we propose Token-Guided Dual Transformer (TGDT) for efficient image-text retrieval. A quick glimpse about the analysis of the accuracy and efficiency of typical approaches is illustrated in Fig.~\ref{fig:speed_acc}. Previous methods such as course-grained and fine-grained retrieval methods could hardly meet the balance of good accuracy and efficiency, as the course-grained methods merely receive high efficiency and the fine-grained retrieval just possess high accuracy.
However, the proposed TGDT-G and TGDT-GL demonstrate both the effectiveness and efficiency for image-text retrieval. In particular, TGDT-GL matches the performance of the recent state-of-the-art vision-language pre-training method VinVL~\cite{VinVL} but significantly reduces the inference time from hundreds of minutes to dozens of seconds.

The main idea which outlines the proposed TGDT is illustrated in Fig.~\ref{fig:introduction}. For each image or text sample, a Transformer based encoder is designed to simultaneously learn a global feature of the overall sample and local features of regional tokens. The compact global feature describes an image and a sentence from a holistic perspective, and local features to automatically aligned to establish a connection between image regions and text words.
To learn a common semantic space shared by the textual and visual inputs, homogeneous network structures are constructed for image and text modalities, respectively. 
Moreover, a novel Consistent Multimodal Contrastive (CMC) loss is proposed to ensure the consistency of global distances between image-text pairs. The CMC loss consists of two central parts which perform contrastive learning between different modalities and within the same modality, respectively. These two losses act cooperatively to ensure the intra- and inter-modal semantic consistencies between images and texts in the common embedding space.
Finally, different inference strategies are exploited based on learned global and local representations, and a fast yet effective two-stage inference method is presented for image-text retrieval.

In summary, the main contributions are listed as follows:
\begin{itemize}
	\item We propose a token-guided dual transformer architecture for image-text retrieval which beneficially leverages the advantages of both coarse- and fine-grained retrieval approaches.
	\item We introduce a consistent multimodal contrastive loss which could guarantee the separation consistency of distances of both the intra- and inter-modal unpaired samples in the common latent space.
	\item We present an effective and efficient inference strategy by sequentially applying global retrieval and local re-ranking in a two-stage manner.
	\item Our methods achieve state-of-the-art performances on several important benchmarks with both high retrieval accuracy and efficiency.
\end{itemize}

\section{Related Work}
\label{sec:Relwork}
\RED{Image-text retrieval is a fundamental application in scene graph~\cite{chang2021comprehensive}. As zero-shot image-text retrieval is an important setting of vision-language pre-training models, recent approaches of zero-shot learning (e.g.,~\cite{yan2021zeronas,zhang2022tn}) could benefit this area.}
Previous works mainly tackle \RED{cross-modal image-text retrieval} from three directions: 1) The coarse-grained retrieval method directly calculate the global similarity between the input image and full text by mapping the two heterogeneous modalities to a common embedding space; 2) The fine-grained matching method automatically aligns image proposals and text fragments by exploring their fine-grained cross-modal correspondences; 3) The visual language pre-training (VLP) method additionally uses external data or knowledge sources to train VLP models in order to learn better representations. 
\vspace{0.2 em}

\noindent\textbf{Coarse-grained Retrieval Methods\ }
With the development of deep learning, end-to-end image-text retrieval becomes prevalent.
Wang \emph{et al.}~\cite{wang2016learning} used two independent multi-layer perceptrons to process images and texts, and adopted structural features for target optimization.
Zheng \emph{et al.}~\cite{DPC} studied the architecture of two independent CNN networks which process images and text, and used instance loss for target optimization.
Faghri \emph{et al.}~\cite{VSE} proposed a training loss based on hard negative samples mining and triplet sampling.
Recently, Gur \emph{et al.}~\cite{DXR} optimized image and text encoders with a small amount of unstructured external knowledge sources of image-text pairs.
These methods process the global image and text information separately through two independent network branches, and have the advantages of fast reasoning and easy pre-calculation. However, they cannot model the fine-grained interactions between object instances and language tokens, and the retrieval performance is roughly limited, especially for complex images and long sentences.
\vspace{0.2 em}

\noindent\textbf{Fine-grained Retrieval Methods\ }
In recent years, more works~\cite{karpathy2017deep,niu2017hierarchical,SCAN,CAMP,VSRN,IMRAM,TERAN,SGRAF,miech2021thinking,zhang2020deep,li2021memorize} are committed to the alignment between text words and image regions.
Karpathy \emph{et al.}~\cite{karpathy2017deep} extracted features for each image region and text word and aligned them in the common embedded space.
Niu \emph{et al.}~\cite{niu2017hierarchical} emphasized the representation of text, and utilized semantic trees and Recurrent Neural Networks (RNN) to extract text phrase features.
Lee \emph{et al.}~\cite{SCAN} introduced multi-layer cross attention between image regions and text words to learn better alignment features.
Wang \emph{et al.}~\cite{CAMP} transmitted lifting alignment effects through a door-adaptive control information on the basis of cross attention.
Li \emph{et al.}~\cite{VSRN} introduced a graph structure for the image region and applied Graph Convolutional Network (GNN) based architectures to extract features.
Chen \emph{et al.}~\cite{IMRAM} introduced an iterative matching scheme to progressively explore such fine-grained correspondence.
Messina \emph{et al.}~\cite{TERAN} designed two transformer encoders to extract features for image regions and text words, respectively.
Diao \emph{et al.}~\cite{SGRAF} employed the graph structure to fuse similarity information and proposed an attention mechanism to filter unimportant elements.
These methods mainly focus on the alignment between fine-grained components of image and text, and often use cross attention mechanisms which require excessive cross computations between different modalities. Although these methods perform well, they show very slow inference speed and are mostly impractical for realistic applications. 
\vspace{0.2 em}

\noindent\textbf{Vision-Language Pre-training Methods\ }
Inspired by the field of natural language processing, the Vision-Language pre-training model using external knowledge sources has gained increasing concerns.
Works on such models~\cite{CLIP,ALIGN,harold2019visualbert,lu2019vilbert,chen2020uniter,UnicoderVL,Oscar,VinVL,tan2019lxmert,zhou2020unified} showed that they can effectively learn general representations from a large number of image-text pairs. And these models can be fine-tuned on the data of a specific task to further improve performance.
CLIP~\cite{CLIP} and ALIGN~\cite{ALIGN} use a large number (0.4 and 1.8 Billion) of noise text-image pairs to train the dual encoder to generate representative features for each modality.
ViLBERT~\cite{lu2019vilbert}, Unicoder-VL~\cite{UnicoderVL}, Oscar~\cite{Oscar}, VinVL~\cite{VinVL} and other methods use large amounts of data on fine-grained images and text elements to train transformer networks.
These methods require vast amounts of external data and reasonable self-supervised tasks to learn excellent pre-training models in order to improve the model's ability on various specific tasks. 
Our method does not use external data during training.

\section{Method}
\begin{figure*}[t]
	\begin{center}
		\includegraphics[width=0.98\linewidth]{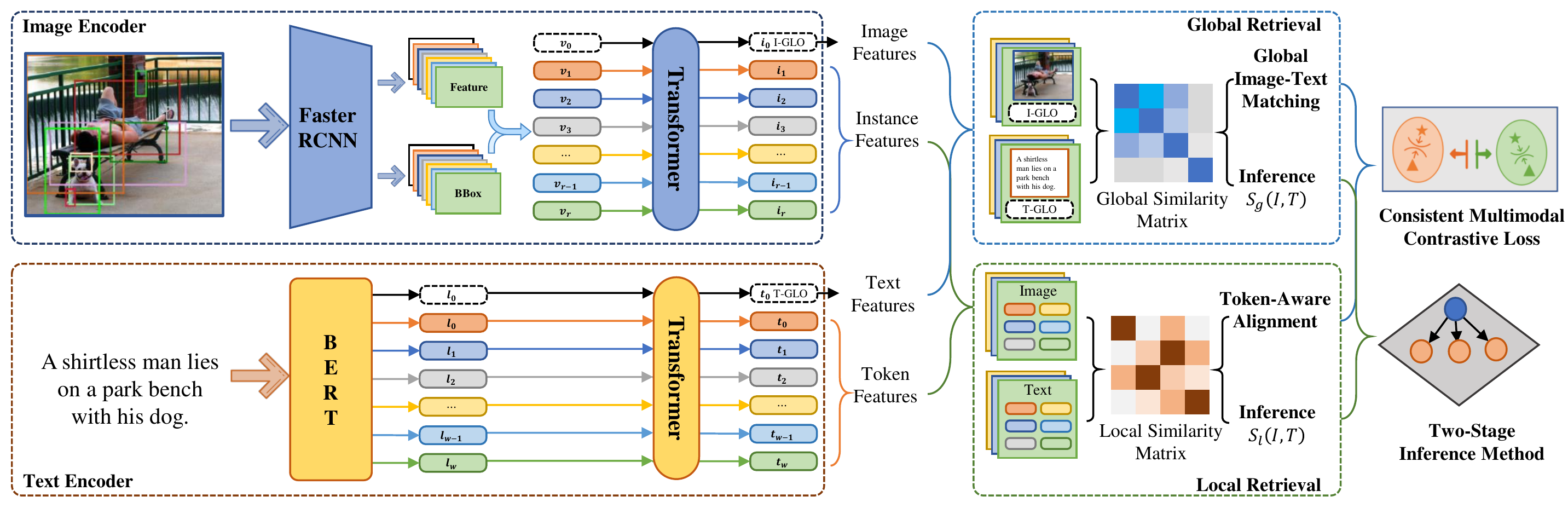}
	\end{center}
\vspace{-1 em}
	\caption{Framework of the proposed Token-Guided Dual Transformer (TGDT) architecture. The image encoder first uses Faster R-CNN to generate locations of image instances as well as their corresponding features and the global image-level features for an input image. These token representations are then passed through the transformer encoder to obtain the cross-modal representations. The text encoder first uses BERT to generate word-level and sentence-level features for each text sample, and then generates the corresponding linguistic representations through another transformer encoder. The global retrieval directly matches the whole image features and the sentence-level text features, while the local retrieval obtains the cross-modal similarity after token-level alignment between image instance features and text word-level features. During training, the proposed Consistent Multimodal Contrastive Training (CMC) loss simultaneously trains the two networks. During inference, a two-stage method based on global and local similarity achieves both accuracy and efficiency.} 
	\label{fig:pipeline}
	\vspace{-0.5 em}
\end{figure*}

In this section, we elaborate details of the proposed Token-Guided Dual Transformer (TGDT) architecture, the framework of which is shown in Fig.~\ref{fig:pipeline}. We first describe the transformer-based cross-modal representation learning. Then, we describe both global and local retrievals for image-text retrieval. Finally, the Consistent Multimodal Contrastive (CMC) training loss is introduced, followed by an efficient inference method.

\subsection{Cross-Modal Representation Learning} 
\label{subsection:Cross-modal Feature Representation}

Cross-modal representation learning involves two different modalities: image and text. To obtain both global and local feature representations simultaneously, we employ the transformer-based architectures for representation learning. Two transformer encoders are designed to process the image and text modalities, respectively, and collaboratively learn the common feature space. 
\vspace{0.2 em}

\noindent\textbf{Image Encoder\ } 
With the development of deep learning in computer vision, convolutional neural networks have become the basis of many visual tasks to extract visual information from images.
Consistent with~\cite{2018Bottom}, and to obtain more descriptive information about visual contents of image regions, we adopt the pre-trained Faster R-CNN~\cite{ren2017faster} as the detector to generate local visual features.

As shown in Fig.~\ref{fig:pipeline}, for a given image, the detector generates $r$ image proposals which describe object instances, as well as the corresponding feature vector $f_i\in \mathbb{R}^d$ and bounding box vector $b_i\in \mathbb{R}^4$ for each proposal. For simplicity, we define the image representation as $V = \{v_0,v_1,\dots,v_r\}$, where $r$ is the number of proposals detected in the image, and $v_i=[f_i,b_i] \in \mathbb{R}^{d+4}$ is the concatenation of $f_i$ and $b_i$. Specifically, when $i = 0$, $v_0$ is the global feature of the image, and the bounding box $b_0=\{0,0,w,h\}$, where $w,h$ is the width and height of the picture.

Recently, transformer-based architectures have performed excellent for both vision and language tasks. The transformer encoder learns representations input tokens, and refers to both global and local information at the same time. We use a transformer architecture to attend features of both image regions and the whole image. This architecture consists of four identical layers of standard transformer encoder, where each layer is composed of a multi-head self-attention mechanism and a fully connected feed-forward operator. 

Let $\mathrm{ITR}(\cdot)$ denote the transformer-based image encoder. Each element in $V$ is used as a token input to the transformer head, and the output is learned image representations:
$$I=\mathrm{ITR}(V)=\{i_0,i_1,\dots,i_r\},$$
where, $i_0$ is the learned global image representation, and $i_1,\dots,i_r$ are the local representations about object instances. 
\vspace{0.2 em}

\noindent\textbf{Text Encoder\ } 
For text representation, the development of natural language processing has given many excellent representation models. Text can be represented at the sentence or word levels. We employ the widely used pre-training model BERT~\cite{devlin2019bert} to extract two levels of text semantic information. For a given sentence of $w$ words, $w+1$ tokens are generated for the entire sentence and as well as the words through BERT. We define the text representation as $L={l_0,l_1,\dots,l_w}$, where $w$ is the number of words in the sentence, $l_0$ is the global representation of the sentence, and $l_1,\dots,l_w$ are local representations of $w$ words. 

Similarly, the transformer-based text encoder, which is denoted as $\mathrm{TTR}(\cdot)$, also has four identical standard transformer layers. Each element in $L$ is regarded as a token input to this transformer head, and the output are text representation features:
$$T=\mathrm{TTR}(L)=\{t_0,t_1,\dots,t_w\},$$
where, $t_0$ is the learned global text representation, and $t_1,\dots,t_r$ are local text representations. 

\subsection{Joint Global and Local Retrievals}
\label{subsection:Global and Local Retrieval Task}
Previous works either use coarse-grained global retrieval or fine-grained local retrieval for image-text retrieval. The proposed TGDT suitably unifies both global and local retrievals under a single framework. The details of two retrieval methods used in our approach are elaborated as follows. 
\vspace{0.2 em}

\noindent\textbf{Global Retrieval\ } 
The global retrieval only uses global features of the two modalities for cross-modal retrieval. Let $X=\{x_0,x_1,\dots,x_{n_x}\}$, $Y=\{y_0,y_1,\dots,y_{n_y}\}$ be the learned representations of two samples with different modalities. For example, $X$ is the output features of the image encoder of one sample, and $Y$ is the output features of the text encoder of another sample. Particularly, $x_0$ and $y_0$ represent the global features of the two samples in the common feature space. The cosine similarity is used to measure the similarity between the two samples. For any two samples, which are represented by $X$ and $Y$, respectively, the global cross-modal similarity is defined as:
\begin{equation}
	S_g(X,Y)=\frac{x_0^T\times y_0}{||x_0||\times||y_0||},
\end{equation}
where $S_g(X,Y)$ only depends on the global features.

An important advantage of global retrieval is that two global features can be calculated independently, and there is no crossover between two samples. Thus, the global retrieval calculation enjoys high speeds, and global features of all the image and text samples can be pre-calculated and stored in a memory to avoid redundant calculations.
\vspace{0.2 em}

\noindent\textbf{Local Retrieval\ } 
The local retrieval fully utilize local features of the two modalities for retrieval. It calculates the similarity by aligning local elements between the two samples. Let $X=\{x_1,\dots,x_{n_x}\}$ and $Y=\{y_1,\dots,y_{n_y}\}$ be the local features in the common feature space of image and text modalities, respectively. We align samples by calculating the cosine similarity between each element: 
\begin{equation}
	M_{ij}(X,Y)=\frac{x_i^T\times y_j}{||x_i||\times||y_j||}.
\end{equation}
$M_{ij}(X,Y)$ represents the similarity between elements $x_i$ and $y_i$, and the two elements from different modalities with the largest similarity are aligned. 

Assume that $X$ is the local image features of one sample and $Y$ is the local text features of another sample, the local similarity is defined as:
\begin{equation}
	S_l(X,Y)=\frac{1}{n_y}\sum_{j\in [1,n_y]}\max_{i\in [1,n_x]} M_{ij}(X,Y),
\end{equation}
where $n_y$ denotes the number of words for the sample represented by $Y$. 
Specifically, $\max_{i\in [1,n_x]} M_{ij}(X,Y)$ finds the most matching element in $X$ for each element in $Y$. Finally, the mean value of the best matching similarities of all elements in $Y$ is taken as the local similarity between the two samples.

The advantage of local similarity is that more refined features achieve better retrieval performance. But the crossover between image and text increases the amount of calculation during retrieval, especially for the cross-attention based and highly entangled approaches which require inefficient fine-grained alignment between words and image regions.


\subsection{Consistent Multimodal Contrastive Loss}
\label{subsection:Intra and Inter Modal Loss}

While training the neural network for image-text retrieval, it is reasonable to ensure the semantic consistency of the image and text samples in the common embedding space. The triplet ranking loss based on multimodal contrastive learning was widely used in previous work~\cite{VSE,IMRAM,TERAN,wang2018learning,wu2017sampling}. The objective is to reduce the distance between matched image-text pairs and pull apart unmatched image-text pairs. However, this loss only considers the similarity between modalities but neglects the relationship between samples of the same modality. In order to ensure the global and local semantic consistency of the samples in the embedding space, we propose a Consistent Multimodal Contrastive (CMC) loss which combines both intra-modal and inter-modal ranking losses during training.

Let $(I,T)$ be a matched image-text pair, where $I$ and $T$ denote the image and text, respectively. 
\RED{We can get the hard negative sample pairs using either the image or the text as the anchor point. For example, with the image $I$ as the anchor point, we can find the hard negative image sample $I_{v^-}$ as well as its corresponding text sample $T_{v^-}$. In this way, two hard negative pairs $(I_{v^-},T_{v^-})$ and $(I_{l^-},T_{l^-})$ are obtained, where $v^-$ and $l^-$ are the indexes of hard negative samples for the image and text modals, respectively.}

\vspace{0.2 em}

\noindent\textbf{Multimodal Contrastive Loss\ } For image-text retrieval, the triplet ranking loss has been widely used in previous works~\cite{wang2018learning,wu2017sampling}. This loss is formally defined as:
\begin{equation}
	\begin{split}
		L_{\text{r}}=&\max(0,\delta-S(I,T)+S(I,T_{l^-}))\\
		+&\max(0,\delta-S(I,T)+S(I_{v^-},T)),
	\end{split}
\end{equation}
where, $S(\cdot)$ is the similarity function, and $\delta$ is a margin hyperparameter which forces the cross-modal distances between both negative pairs $(I, T_{l^-})$ and $(I_{v^-},T)$ are greater than the positive pair $(I,T)$ by a margin of $\delta$.

\begin{figure}[t]
	\begin{center}
		\includegraphics[width=0.8\linewidth]{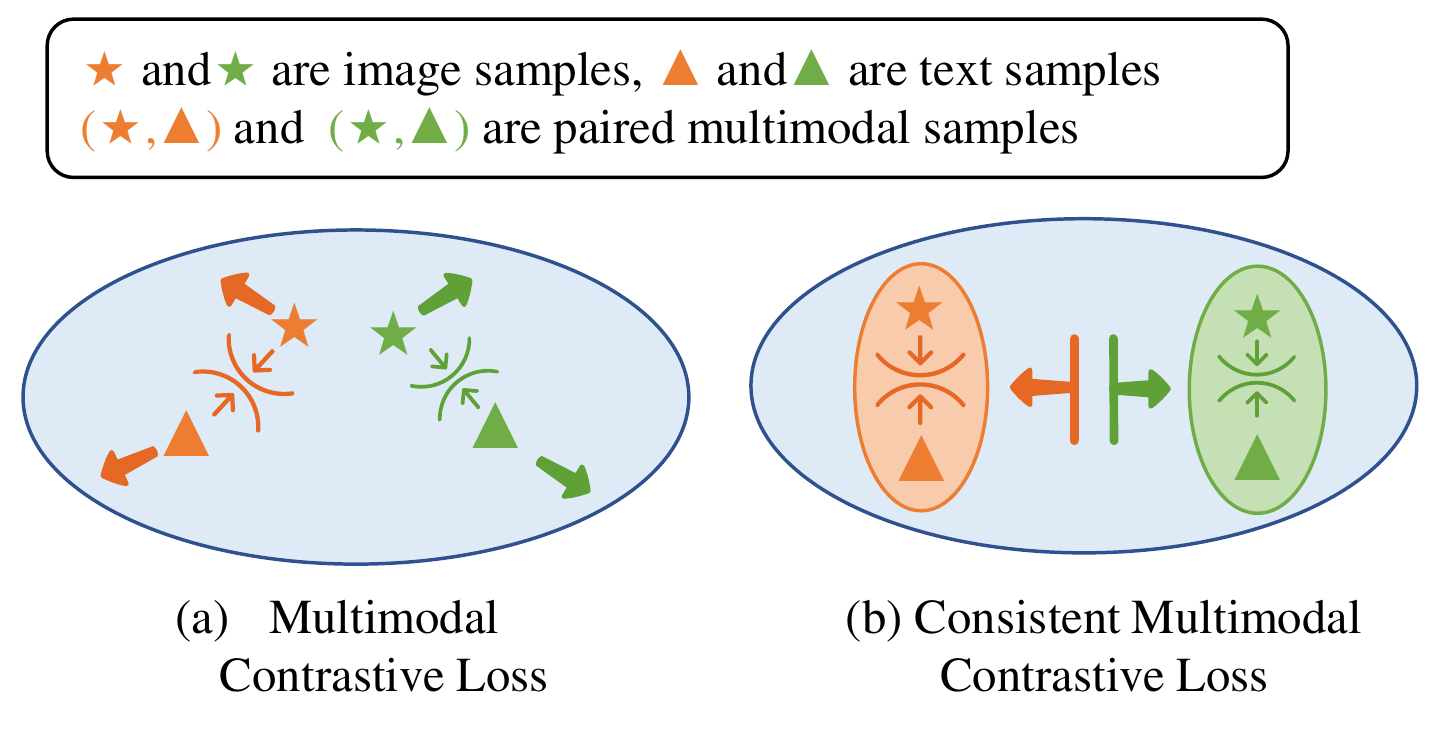}
	\end{center}
\vspace{-1 em}
	\caption{Schematic diagram of two types of loss. (a) Previous multimodal contrastive loss cannot control the distance between samples of different modalities, as it only reduces distances between matched samples and increase distances between unmatched samples. (b) The proposed Consistent Multimodal Contrastive (CMC) loss can ensure the consistency of the distance between each sample regardless of the same or different modalities.} 
	\label{fig:I2loss}
\vspace{-0.5 em}
\end{figure}

As shown in Fig.~\ref{fig:I2loss}(a), the previous multimodal contrastive loss $L_{\text{r}}$ only perform contrastive learning between different modalities, and can be considered as the inter-modal loss. Although this loss pulls apart unmatched samples with different modalities, it lacks the constraint between samples of the same modality. As a result, the distance between samples of the same modality that do not match may be very close.
\medskip

\noindent\textbf{Consistent Multimodal Contrastive Loss\ } The basic intuition behind the proposed Consistent Multimodal Contrastive (CMC) loss is to consider the matched image-text pair as a compact sample and try to ensure the consistency of the global distance between different pairs. Given the three image-text pairs $(I,T)$, $(I_{v^-},T_{v^-})$ and $(I_{l^-},T_{l^-})$ mentioned above, we first define the contrastive loss within the same modality, whose definition is as follows:
\begin{equation}
	\begin{split}
		L_{\text{a}}=&\max(0,|S(I,I_{l^-})-S(T,T_{l^-})|-\sigma)\\
		+&\max(0,|S(I,I_{v^-})-S(T,T_{v^-})|-\sigma),
	\end{split}
	\label{eq:intra}
\end{equation}
where, $S(\cdot)$ is the similarity function, and $\sigma$ is a slack variable, which guarantees that the distance between samples of different modalities can have a particular gap and may not be entirely consistent for the purpose of relaxing the constraint on distances between samples.

Mathematically, the intra-modal loss defined in Eq.~\ref{eq:intra} ensures that $S(I,I_{l^-})$ has a small gap with $S(T,T_{l^-})$, and that $S(I,I_{v^-})$ has a small gap with $S(T,T_{v^-})$. As shown in Fig.~\ref{fig:I2loss}~(b), the primary purpose of the intra-modal loss is to ensure that the distance between two pairs of samples (marked with orange and green colors) of the same modality is consistent. The goal is to make the distance between the orange image sample and the green image sample close to that between the orange text sample and the green text sample.

In view of the above discussion, the proposed CMC loss can be directly obtained by adding the inter-modal loss and the intra-modal loss:
\begin{equation}
	L_{\text{cmc}}=L_{\text{r}}+L_{\text{a}}.
\end{equation}

As shown in Fig.~\ref{fig:I2loss}(b), the proposed loss $L_{\text{cmc}}$ can ensure that the same-modal distance and cross-modal distance of two pairs of samples are consistent, and that the connection between the same pair of matched samples is close. On the one hand, $L_{\text{r}}$ controls the distance between the samples. On the other hand, $L_{\text{a}}$ guarantees the consistency of the distance between the matched samples. Therefore, combining the two losses can ensure the consistency of both local and global similarities between multimodal samples.

\subsection{Two-Stage Inference Method}
\label{subsection:Pipeline}

As mentioned above, the two independent global or local retrieval tasks have their own advantages but cannot achieve both high accuracy and efficiency at the same time. In addition, global retrieval based approaches lacks an understanding of local semantic information, and local retrieval based approaches usually neglect the global description. These reasons limit the performance of both types of models. 
In the proposed TGDT, we mix both global and local retrieval in order to address the above problems and retain advantages of the two retrieval tasks. Details of the training and inference stages of our approach are presented below.
\vspace{0.2 em}

\noindent\textbf{Training\ }Given the defined both global and local cross-modal similarities $S_g(\cdot)$ and $S_l(\cdot)$ in subsection~\ref{subsection:Global and Local Retrieval Task}, as well as the proposed CMC loss function $L_{\text{cmc}}$ in subsection~\ref{subsection:Intra and Inter Modal Loss}, the final training loss of our approach is as follows:
\begin{equation}
	L_{\text{TGDT}}=L_{\text{cmc}}(S_g)+L_{\text{cmc}}(S_l),
\end{equation}
where $L_{\text{cmc}}(S_g)$ and $L_{\text{cmc}}(S_l)$ are the proposed CMC losses when $S(\cdot)$ is $S_g(\cdot)$ and $S_l(\cdot)$, respectively.

The training loss combines both global and local retrieval tasks and jointly learns both global and local representations of two modalities (image and text) in an end-to-end trainable network. The multi-task loss limits the parameter search space so that the model could produce better representations. Subsequent experiments verify that simultaneously training the two tasks could improve the performances of both two tasks when they are trained individually. 
\vspace{0.2 em}

\noindent\textbf{Inference\ }
During inference, we have three ways to calculate the similarity between two samples: 1) Use only global features to calculate the global similarity $S_g(X,Y)$; 2) Use only local features to calculate local similarity $S_l(X,Y)$; 3) Use two features at the same time to obtain mixed similarity:
\begin{equation}
	S_{gl}(X,Y)=(1-\theta)S_g(X,Y)+\theta S_l(X,Y),
\end{equation}
where $\theta$ is the hyperparameter to adjust the ratio of the two similarities. Retrieval accuracies and inference speeds of the three variations are compared in subsequent experiments.

Global retrieval often has fast inference speed but limited retrieval accuracy, and local retrieval often has high retrieval accuracy but slow inference speed. To balance retrieval accuracy and speed, we design a two-stage inference method which elegantly uses a combination of both global and local retrievals for inference. First, global retrieval is used to obtain the top $K$ candidate samples quickly. Then, mixed similarity based on both global and local retrievals is applied to re-rank only the top $K$ samples.

Taking image$\rightarrow$text retrieval as an example, the inference process given an input image query $I_i$ and a text dataset $\mathcal{T}$ containing $n$ text samples is described below. First, we use global retrieval to quickly calculate the similarity between the query image $I_i$ and each text $T_j$ based on $\{S_g(I_i,T_j)|T_j\in\mathcal{T}\}$. Then, the top $K$($K<<n$) text samples $\mathcal{T_K}$ with the greatest similarity are selected as candidate samples, and the mixed similarity $\{S_{gl}(I_i,T_j)|T_j\in\mathcal{T_K}\}$ is calculated to obtain more accurate similarities between the query image and candidate samples to obtain the final ranking.
\RED{The idea of this inference strategy is also named as geometric verification~\cite{cao2020unifying,noh2017large} in image retrieval, which first selects the most similar images using global features, then re-ranks top results using local features. We first apply this strategy for cross-modal retrieval.}

\section{Experiment}

In this section, we show the experiments of our method on two important benchmark datasets: Flickr30K~\cite{young2014F30K} and COCO~\cite{lin2014COCO}. 
Extensive and comprehensive ablation studies are provided to verify the effectiveness of the constructed modules of the proposed method. 

\subsection{Dataset and Evaluation Metric}
\noindent \textbf{Dataset\ } 
The \textbf{Flickr30K} dataset contains 31,783 images, and each image has 5 corresponding texts. Images are collected from the Flickr website, and each image has 5 text descriptions. 
Consistent with~\cite{VSE,IMRAM,TERAN}, we divide the data set into 29,783 images for training, 1,000 images for validation, and 1,000 images for testing. 

The \textbf{COCO} dataset is one of the most popular and important benchmarks for image and sentence matching tasks. It has 123,287 images, which are divided into 113,287 images for training, 5,000 images for verification, and 5,000 images for testing. Every image is matched with 5 corresponding texts, and the average length of an text is 8.7. Consist with previous practices~\cite{VSE,CAMP,TERAN,UnicoderVL}, we provide experimental results for \textbf{COCO 5K} and \textbf{COCO 1K}, respectively. The COCO 5K uses a complete 5,000 images for testing, while the COCO 1K uses 1,000 images for five independent tests.
\vspace{0.2 em}

\noindent \textbf{Evaluation Metrics\ }
For image-text retrieval tasks, rank at top-K (R@K) is a widely adopted as the evaluation metrics, whose definition is the proportion of ground truth contained in the top-K samples of the retrieval results. By convention, R@1, R@5, and R@10 are used to quantitatively measure experimental performance. In order to measure the method's efficiency, we also count the whole inference time for all samples in the test dataset. 
\vspace{0.2 em}

\noindent \textbf{Implementation Details\ } 
For the image input, we use the object proposals provided by~\cite{2018Bottom}, which selects the top 36 region proposals with the highest confidence scores, and describes each object proposal with a 2048-dimensional bottom-up feature vector.
For the text input, we use the BERT model~\cite{BERTPre} which is pre-trained on the mask language task of English sentences to obtain the 768-dimensional embedding features of the text. 
For transformer encoders of both modalities, the embedding size of self-attention layers is 1024, and the output feature dimension is 2048.
The default margin or slack hyperparameters $\delta=0.2$ and $\sigma=0.3$ in the proposed CMC loss function are 0.2 and 0.3, respectively. 
During training, we use Adam as the optimizer. The initial learning rate, number of training epochs, and batch size are $1e-6$, 30 and 40, respectively. 
During inference, the hyperparameter $\theta$ in the mixed similarity is 0.5, and the number of selected samples $K$ in the two-stage inference is 100.
Detailed analysis on the sensitivity of these hyperparameters are conducted in subsequent experiments.

\begin{table}[t]
	\caption{Comparison with the state-of-the-art on the \textbf{Flickr30K} dataset. The symbols $\mathrm{G}$ and $\mathrm{L}$ denote the coarse-grained and fine-grained retrieval methods, respectively.}
	\vspace{-2 em}
	\begin{center}
		\resizebox{1\linewidth}{!}{
			\begin{tabular}{@{}lcccccccc}
				\toprule
				\multirow{2}{*}{Method} & \multirow{2}{*}{G} &  \multirow{2}{*}{L} 
				& \multicolumn{3}{c}{Text → Image}  
				& \multicolumn{3}{c}{Image → Text}  \\
				\cmidrule(l){4-9}
				&  & & R@1           & R@5           & R@10          & R@1           & R@5           & R@10  \\ 
				\midrule
				DPC~\cite{DPC}  & \checkmark  &     & 39.1          & 69.2          & 69.2          & 55.6          & 81.9          & 89.5    \\
				VSE++~\cite{VSE}   & \checkmark  &             & 39.6          & 69.6          & 79.5          & 52.9          & 79.1          & 87.2       \\
				DXR~\cite{DXR} & \checkmark &             & 50.6          & 78.8          & 86.7          & 65.1          & 87.3          & 92.6      \\
				\RED{MFM~\cite{Ma2020MatchingIA}} & \checkmark & &  38.2 & 70.1 & 80.2 & 50.2 & 78.1 & 86.7 \\
				\midrule
				CAMP~\cite{CAMP}  & & \checkmark  & 51.5          & 77.1          & 85.3          & 68.1          & 89.7          & 95.2    \\
				SCAN~\cite{SCAN}  & & \checkmark  & 48.6          & 77.7          & 85.2          & 67.4          & 90.3          & 95.8   \\
				\RED{SCAN-L\textsubscript{Rel}~\cite{wei2021universal}} & & \checkmark &  51.0 & 76.8 & 84.4 & 66.5 & 91.1 & 95.8 \\
				\RED{DSRAN~\cite{wen2020learning}} & & \checkmark & 57.3 & 84.8 & 90.9 & 75.3 & 94.4 & 97.6\\
				\RED{SMAN~\cite{ji2020sman}} &  & \checkmark  & 43.4 & 73.7 & 83.4 & 57.3 & 85.3 & 92.2 \\
				\RED{CAMERA~\cite{qu2020context}} & &  \checkmark & 60.3 & 85.9 & 91.7 & 78.0 & 95.1 & 97.9  \\
				VSRN~\cite{VSRN} & & \checkmark  & 54.7          & 81.8          & 88.2          & 71.3          & 90.6          & 96.0   \\
				IMRAM~\cite{IMRAM} & & \checkmark  & 53.9          & 79.4          & 87.2          & 74.1          & 93.0          & 96.6     \\
				CRGN~\cite{zhang2020deep} & & \checkmark & 50.3 & 77.7 & 85.2 & 70.5 & 91.2 & 94.9 \\
				SGRAF~\cite{SGRAF} & & \checkmark  & 58.5          & 83.0          & 88.8          & 77.8 & 94.1 & 97.4  \\
				MEMBER~\cite{li2021memorize} & & \checkmark  & 59.5 & 84.8 & 91.0 & 77.5 & 94.7 & 97.3 \\
				TERAN~\cite{TERAN} & & \checkmark & 56.5          & 81.2          & 88.2          & 70.8          & 90.9          & 95.5     \\
				\RED{NCR~\cite{huang2021learning}} & &  \checkmark & 59.6 & 84.4 & 89.9 & 77.3 &  94.0 & 97.5 \\
				\RED{DCPG~\cite{yan2021discrete}}  & &  \checkmark & \textbf{62.2} & \textbf{89.3} & \textbf{93.8} & \textbf{82.8} & \textbf{95.9} & \textbf{97.9} \\
				\RED{CGMN~\cite{cheng2022cross}}  & &  \checkmark & 59.9 & 85.1 & 90.6 & 77.9 & 93.8 & 96.8 \\
				\RED{UARDA~\cite{zhang2022unified}} & &  \checkmark & 57.8 & 82.9 & 89.2 & 77.8 & 95.0 & 97.6 \\
				\midrule
				\textbf{TGDT-G} & \checkmark &       & 55.6 & 83.1 & 89.4 & 70.3 & 91.4 & 95.5 \\ 
				\textbf{TGDT-L} & & \checkmark  & 61.3 & 86.0 & 91.4 & 76.8          & 93.2          & 96.4 \\ 
				\textbf{TGDT-GL} &\checkmark  & \checkmark  & \textbf{66.7} & \textbf{92.2} & \textbf{97.0} & \textbf{79.6} & \textbf{96.9} & \textbf{99.0} \\
				\bottomrule
			\end{tabular}
		}
	\end{center}
 \vspace{-1 em}
	\label{tab:Flickr30K}
\end{table}

\begin{table}[t]
	\caption{Comparison with the state-of-the-art on the \textbf{COCO 1K} dataset.}
	\vspace{-2 em}
	\begin{center}
		\resizebox{1\linewidth}{!}{
			\begin{tabular}{@{}lcccccccc}
				\toprule
				\multirow{2}{*}{Method} & \multirow{2}{*}{G} &  \multirow{2}{*}{L}
				& \multicolumn{3}{c}{Text → Image}  
				& \multicolumn{3}{c}{Image → Text}  \\
				\cmidrule(l){4-9} 
				& & & R@1           & R@5           & R@10          & R@1           & R@5           & R@10  \\ 
				\midrule
				DPC~\cite{DPC} & \checkmark  &  & 47.1          & 79.9          & 90.0          & 65.6          & 89.8          & 95.5           \\
				VSE++~\cite{VSE} & \checkmark  & & 52.0          & 83.1          & 92.0          & 64.6          & 89.1          & 95.7       \\
				DXR~\cite{DXR} & \checkmark  &  & 56.8          & 88.2          & 94.9          & 67.0          & 93.0          & 97.6    \\
				\RED{MFM~\cite{Ma2020MatchingIA}} & \checkmark & & 47.7 & 81.0 & 90.9 & 58.9 & 86.3 & 92.4\\
				\midrule
				CAMP~\cite{CAMP} &  & \checkmark & 58.5          & 87.9          & 95.0          & 72.3          & 94.8          & 98.3  \\
				SCAN~\cite{SCAN} &  & \checkmark & 58.8          & 88.4          & 94.8          & 72.7          & 94.8          & 98.4    \\
				\RED{SCAN-L\textsubscript{Rel}~\cite{wei2021universal}} & & \checkmark & 58.4 & 87.3 & 93.7 & 74.5 & 94.9 & 98.1 \\
				\RED{DSRAN~\cite{wen2020learning}} & & \checkmark & 62.9 & 89.9 & 95.3 & 77.1 & 95.3 & 98.1\\
				\RED{SMAN~\cite{ji2020sman}} &  & \checkmark  & 58.5 & 87.4 & 93.5 & 68.4 & 91.3 & 96.6 \\ 
				\RED{CAMERA~\cite{qu2020context}} & &  \checkmark  & 63.4 & 90.9 & 95.8 & 77.5 & 96.3 & 98.8 \\
				VSRN~\cite{VSRN} &  & \checkmark &  62.8          & 89.7          & 95.0          & 76.2          & 94.8          & 98.2  \\
				CRGN~\cite{zhang2020deep} & & \checkmark & 60.1 & 88.9 & 94.5 & 73.8 & 95.6 & 98.5 \\
				IMRAM~\cite{IMRAM} &  & \checkmark & 61.7          & 89.1          & 95.0          & 76.7          & 95.6          & 98.5    \\
				SGRAF~\cite{SGRAF} &  & \checkmark & 63.2          & 90.7          & 96.1          & \textbf{79.6}          & 96.2          & 98.5  \\
				MEMBER~\cite{li2021memorize} & & \checkmark  & 63.7 & 90.7 & 95.6 & 78.5 & 96.8 & 98.5\\
				TERAN~\cite{TERAN} &  & \checkmark &  \textbf{65.0}          & \textbf{91.2}          & \textbf{96.4}          & 77.7          & 95.9          & 98.6         \\
				\RED{NCR~\cite{huang2021learning}} & &  \checkmark & 63.3 & 90.4 & 95.8 & 78.7 & 95.8 & 98.5 \\
				\RED{DCPG~\cite{yan2021discrete}} & &  \checkmark & 63.9 & 88.9 & 95.6 & 84.0 & 95.8 & 97.8 \\
				\RED{CGMN~\cite{cheng2022cross}}  & &  \checkmark & 63.8 & 90.7 & 95.7 & 76.8 & 95.4 & 98.3 \\
				\RED{UARDA~\cite{zhang2022unified}} & &  \checkmark & 63.9 & 90.7 & 96.2 & 78.6 & \textbf{96.5} & \textbf{98.9} \\
				\midrule
				\textbf{TGDT-G} & \checkmark  &  & 61.3 & 89.8 & 95.3 & 73.8 & 94.6 & 98.2  \\ 
				\textbf{TGDT-L} &  & \checkmark &65.4 & 91.8 & 96.5 & 78.5 & 96.4 & 98.9 \\ 
				\textbf{TGDT-GL} & \checkmark & \checkmark & \textbf{66.7} & \textbf{92.2} & \textbf{97.0} & \textbf{79.6} & \textbf{96.9} & \textbf{99.0} \\
				\bottomrule
			\end{tabular}
		}
	\vspace{-1 em}
	\end{center}
	\label{tab:COCO_1K}
\end{table}

\begin{table}[t]
	\caption{Comparison with the state-of-the-art on the \textbf{COCO 5K} dataset.}
	\vspace{-2 em}
	\begin{center}
		\resizebox{1\linewidth}{!}{
			\begin{tabular}{@{}lcccccccc}
				\toprule
				\multirow{2}{*}{Method} & \multirow{2}{*}{G} &  \multirow{2}{*}{L}
				& \multicolumn{3}{c}{Text → Image}  
				& \multicolumn{3}{c}{Image → Text}  \\
				\cmidrule(l){4-9} 
				& & & R@1           & R@5           & R@10          & R@1           & R@5           & R@10  \\ 
				\midrule
				DPC~\cite{DPC} & \checkmark  &  & 25.3          & 53.4          & 66.4          & 41.2          & 70.5          & 81.1          \\
				VSE++~\cite{VSE} & \checkmark  & & 30.3          & 59.1          & 72.4          & 41.3          & 69.2          & 81.2          \\
				DXR~\cite{DXR} & \checkmark  &  & 33.9          & 64.9          & 77.4          & 44.9          & 75.2          & 84.7          \\
				\midrule
				CAMP~\cite{CAMP} &  & \checkmark & 39.0          & 68.9          & 80.2          & 50.1          & 82.1          & 89.7          \\
				SCAN~\cite{SCAN} &  & \checkmark & 38.6          & 69.3          & 80.4          & 50.4          & 82.2          & 90.0          \\
				\RED{SCAN-L\textsubscript{Rel}~\cite{wei2021universal}} & & \checkmark & 34.4 & 64.2 & 75.9 & 46.9 & 77.7 & 87.6\\
				\RED{DSRAN~\cite{wen2020learning}} & & \checkmark & 40.3 & 70.9 & 81.3 & 53.7 & 82.1 & 89.9 \\
				\RED{CAMERA~\cite{qu2020context}} & &  \checkmark  & 40.5 & 71.7 & 82.5 & 55.1 & 82.9 & 91.2  \\
				VSRN~\cite{VSRN} &  & \checkmark & 40.5          & 70.6          & 81.1          & 53.0          & 81.1          & 89.4          \\
				CRGN~\cite{zhang2020deep} & & \checkmark & 37.4 & 68.0 & 79.5 & 51.2 & 80.6 & 89.7 \\
				IMRAM~\cite{IMRAM} &  & \checkmark & 39.7          & 69.1          & 79.8          & 53.7          & 83.2          & 90.1          \\
				SGRAF~\cite{SGRAF} &  & \checkmark &  41.9          & -             & 81.3          & 57.8 & -             & 91.6          \\
				MEMBER~\cite{li2021memorize} & & \checkmark  & 40.9 & 71.0 & 81.8 & 54.5 & 82.3 & 90.1 \\
				TERAN~\cite{TERAN} &  & \checkmark & 42.6          & 72.5          & 82.9          & 55.6          & 83.9          & 91.6          \\
				\RED{DCPG~\cite{yan2021discrete}} & &  \checkmark & \textbf{46.2} & \textbf{77.8} & \textbf{85.5} & \textbf{68.7} & \textbf{88.7} & \textbf{93.0} \\
				\RED{CGMN~\cite{cheng2022cross}}  & &  \checkmark  & 41.2 & 71.9 & 82.4 & 53.4 & 81.3 & 89.6 \\
				\RED{UARDA~\cite{zhang2022unified}} & &  \checkmark & 40.6 & 69.5 & 80.9 & 56.2 & 83.8 & 91.3 \\
				\midrule
				\textbf{TGDT-G} & \checkmark  &  & 38.2 & 69.1 & 80.6 & 50.1 & 78.8 & 88.1 \\ 
				\textbf{TGDT-L} &  & \checkmark & 43.3 & 73.5 & 83.3 & 57.5          & 84.8 & 91.6 \\ 
				\textbf{TGDT-GL} & \checkmark & \checkmark &\textbf{45.1} & \textbf{75.2} & \textbf{85.1} & \textbf{59.3}       & \textbf{85.7} & \textbf{92.8} \\
				\bottomrule
			\end{tabular}
		}
	\end{center}
\vspace{-1 em}
	\label{tab:COCO_5K}
\end{table}

\subsection{Results of Image-Text Retrieval} \label{subsection:results_1}
We evaluate our methods on Flickr30K, COCO 1K, and COCO 5K, and compare results with those of state-of-the-art approaches. As our methods do not use external data during
training, we do not compare them with the visual language pre-training (VLP) methods in this subsection.

\noindent \textbf{Comparisons on Flickr30K\ } 
The results on the Flickr30K dataset are summarized in Tab.~\ref{tab:Flickr30K}. For the sake of simplicity, $\mathrm{G}$ denotes the coarse-grained retrieval method which only uses global representations of image and text during inference, $\mathrm{L}$ denotes the fine-grained retrieval method which depends on local representations of image regions or word tokens for inference. TGDT-G and TGDT-L are the coarse-grained and fine-grained variants of our approach, respectively, and only utilize global or local representations in the inference process. TGDT-GL is the proposed approach that adopts the two-stage inference using both global and local representations.

For the coarse-grained category, TGDT-G dramatically outperforms the recent state-of-the-art approaches for both tasks of text-to-image retrieval and image-to-text retrieval. 
For the fine-grained category, TGDT-L also exceeds \RED{most recent} state-of-the-art methods by considerable margins. Moreover, TGDT-GL consistently outperforms both TGDT-G and TGDT-L, \RED{which demonstrates the complementarity of global representations and local representations.} 

\vspace{0.2 em}
\noindent \textbf{Comparisons on COCO\ } The results on the COCO 1K and COCO 5K are shown in Tab.~\ref{tab:COCO_1K} and Tab.~\ref{tab:COCO_5K}, respectively. 
\RED{TGDT-G achieves the best results among all the coarse-grained methods on both the COCO 1K and COCO 5K datasets, while TGDT-GL significantly outperforms most state-of-the-art fine-grained methods under different settings.}
The results demonstrate the effectiveness of both modules of global retrieval and local retrieval in the proposed TGDT architecture.

\begin{table}[ht]
	\begin{center}
		\caption{Comparison with vision-language pre-training models on the \textbf{Flickr30K} dataset. }
		\vspace{-1 em}
		\resizebox{1\linewidth}{!}{
			\begin{tabular}{lcccccc}
				\toprule
				\multicolumn{1}{l}{\multirow{3}{*}{Method}} & \multicolumn{6}{c}{Flickr30K}                                                            \\ \cmidrule(l){2-7} 
				\multicolumn{1}{c}{}                        & \multicolumn{3}{c}{Text→Image} & \multicolumn{3}{c}{Image→Text}  \\
				\multicolumn{1}{c}{}                        & R@1      & R@5      & R@10     & R@1      & R@5      & R@10    \\ \midrule
				CLIP~\cite{CLIP}                                        & 68.7     & 90.6     & 95.2     & 88.0     & 98.7     & 99.4      \\
				ALIGN~\cite{ALIGN}                                       & 75.7     & 93.8     & 96.8     & 88.6     & 98.7     & 99.7         \\
				Unicoder-VL~\cite{UnicoderVL}                                 & 71.5     & 90.9     & 94.9     & 86.2     & 96.3     & 99.0     \\
				Oscar~\cite{Oscar}                                       & 75.9     & 93.3     & 96.6     & 88.5     & 98.5     & 99.2         \\
				\RED{UNITER~\cite{chen2020uniter}} & 72.5 & 92.4 & 96.1 & 85.9 & 97.1 & 98.8 \\
				\RED{UNITER-IAIS~\cite{ren2021learning}} & 76.9 & 93.3 & 95.7 & 88.3 & 98.4 & 99.4 \\
				\RED{VILLA~\cite{gan2020large}} & 74.7 & 92.9 & 95.8 & 86.6 & 97.9 & 99.2 \\
				ERNIE-ViL-B~\cite{yu2021ernie} & 74.4 & 92.7 & 95.9 & 86.7 & 97.8 & 99.0 \\
				VSEinfty~\cite{chen2021learning} & 74.2 & 93.7 & 96.8 & 88.4 & 98.3 & 99.5 \\
				SOHO~\cite{huang2021seeing} & 72.5 & 92.7 & 96.1 & 86.5 & 98.1 & 99.3 \\
				SAVPVLP~\cite{xue2021probing} & 73.5 & 93.1 & 96.4 & 87.0 & 98.4 & 99.5 \\
				VinVL~\cite{VinVL}  & 76.4     & 93.9     & 97.1     & 88.8   & 98.7     & 99.3    \\ 
				ViLT-B~\cite{kim2021vilt} & 64.4 & 88.7 & 93.8 & 83.5 & 96.7 & 98.6 \\ 
				\RED{Pixel-Bert~\cite{huang2020pixel}} & 59.8 & 85.5 & 91.6 & 75.7 & 94.7 & 97.1 \\
				\RED{ERNIE-ViL~\cite{yu2021ernie}} & 74.4 & 92.7 & 95.9 & 86.7 & 97.8 & 99.0 \\
				\RED{LightningDOT~\cite{sun2021lightningdot}} & 69.9 & 91.1 & 95.2 & 83.9 & 97.2 & 98.6 \\
				\RED{ALBEF~\cite{li2021align}} & 82.8 & 96.7 & 98.4 & 94.3 & 99.4 & 99.8 \\
				\RED{TCL~\cite{yang2022vision}} & \textbf{84.0} & \textbf{96.7} & \textbf{98.5} & \textbf{94.9} & \textbf{99.5} & \textbf{99.8} \\
				ViSTA-B~\cite{cheng2022vista} & 68.9 & 91.1 & 95.1 & 84.8 & 97.4 & 99.0 \\
				\RED{CP BERT~\cite{yu2022cross}} & 69.1 & 89.8 & 94.1 & 83.5 & 96.0 & 98.0 \\
				\RED{Knowledge-CLIP~\cite{pancontrastive}} & 75.7 & 94.4 & 96.8 & 89.2 & 98.9 & 99.4 \\
				\midrule
				\textbf{TGDT-G} & 68.3     & 90.5     & 96.4     & 82.5     & 95.4     & 98.8       \\
				\textbf{TGDT-L}  & 76.7     & 94.2     & 97.2     & 88.6     & 98.5     & 99.2      \\
				\textbf{TGDT-GL}  & \textbf{77.4}     & \textbf{94.4}     & \textbf{97.3}     & \textbf{89.2}     & \textbf{98.7}     & \textbf{99.3}      \\ \bottomrule
			\end{tabular}
		}
	\end{center}
\vspace{-1 em}
	\label{tab:F30Koscar}
\end{table}

\begin{table*}[ht]
	\caption{Comparison with vision-language pre-training models on the \textbf{COCO 1K} and \textbf{COCO 5K} datasets.}
	\begin{center}
		\resizebox{0.91\linewidth}{!}{
			\begin{tabular}{lcccccc|cccccc}
				\toprule
				\multicolumn{1}{l}{\multirow{3}{*}{Method}} & \multicolumn{6}{c|}{COCO 1K}                                     & \multicolumn{6}{c}{COCO 5K}                                     \\ \cmidrule(l){2-13} 
				\multicolumn{1}{c}{}                        & \multicolumn{3}{c}{Text→Image} & \multicolumn{3}{c|}{Image→Text} & \multicolumn{3}{c}{Text→Image} & \multicolumn{3}{c}{Image→Text} \\
				\multicolumn{1}{c}{}                        & R@1      & R@5      & R@10     & R@1       & R@5      & R@10     & R@1      & R@5      & R@10     & R@1      & R@5      & R@10     \\ \midrule
				CLIP~\cite{CLIP}  & -- & -- & -- & -- & -- & -- & 37.8     & 62.4     & 72.2     & 58.4     & 81.5     & 88.1     \\
				ALIGN~\cite{ALIGN}  & -- & -- & -- & -- & -- & -- & 45.6     & 69.8     & 78.6     & 58.6     & 83.0     & 89.7     \\
				Unicoder-VL~\cite{UnicoderVL}   & 69.7     & 93.5     & 97.2     & 84.3      & 97.3     & 99.3     & 46.7     & 76.0     & 85.3     & 62.3     & 87.1     & 92.8     \\
				Oscar~\cite{Oscar}  & 78.2     & 95.8     & 98.3     & 89.8      & 98.8     & 99.7     & 57.5     & 82.8     & 89.8     & 73.5     & 92.2     & 96.0     \\
				\RED{UNITER~\cite{chen2020uniter}} & -- & -- & -- & -- & -- & -- & 50.3 & 78.5 & 87.2 & 64.4 & 87.4 & 93.1 \\
				\RED{UNITER-IAIS~\cite{ren2021learning}} & -- & -- & -- & -- & -- & -- & 53.2 & 80.1 & 87.9 & 67.8 & 89.7 & 94.5 \\
				\RED{VSEinfty~\cite{chen2021learning}} & 72.0 & 93.9 & 97.5 & 84.5 & 98.1 & 99.4  & -- & -- & -- & -- & -- & -- \\
				SOHO~\cite{huang2021seeing} & 73.5 & 94.5 & 97.5 & 85.1 & 97.4 & 99.4 & 50.6 & 78.0  & 86.7 & 66.4 & 88.2 & 93.8 \\
				VinVL~\cite{VinVL} & \textbf{78.8}     & \textbf{96.1}     & \textbf{98.5}    & \textbf{90.8}     & \textbf{99.0}     & \textbf{99.8}     & 58.8     & 83.5     & 90.3  & \textbf{75.4} & \textbf{92.9} & 96.2   \\
				ViLT-B~\cite{kim2021vilt}  & -- & -- & -- & -- & -- & -- & 42.7 & 72.9 & 83.1 & 61.5 & 86.3 & 92.7 \\
				\RED{Pixel-Bert~\cite{huang2020pixel}} & 64.1 & 91.0 & 96.2 &  77.8 & 95.4 & 98.2 &  41.1 & 69.7 & 80.5 &  53.4 & 80.4 & 88.5 \\
				\RED{LightningDOT~\cite{sun2021lightningdot}} & -- & -- & -- & -- & -- & -- &  45.8 & 74.6 & 83.8 & 60.1 & 85.1 & 91.8  \\
				\RED{ALBEF~\cite{li2021align}} &  -- & -- & -- & -- & -- & -- &  56.8 & 81.5 & 89.2 & 73.1 & 91.4 & 96.0 \\
				\RED{TCL~\cite{yang2022vision}} & -- & -- & -- & -- & -- & -- &  \textbf{59.0} & 83.2 & 89.9 & 75.6 & 92.8 & \textbf{96.7} \\
				ViSTA-B~\cite{cheng2022vista} & -- & -- & -- & -- & -- & -- & 47.8 & 75.8 & 84.5 &    63.9 & 87.8 & 93.6\\
				\RED{CP BERT~\cite{yu2022cross}} & 70.9 & 92.5 & 96.6 & 83.3 & 96.9 & 99.4 & 46.8 & 75.8 & 85.0 & 62.9 & 86.7 & 92.7 \\
				\RED{Knowledge-CLIP~\cite{pancontrastive}} & -- & -- & -- & -- & -- & -- & 57.6 & \textbf{83.9} & \textbf{90.4} & 70.2 & 89.2 & 94.4 \\
				\midrule
				\textbf{TGDT-G }   & 69.2     & 93.1     & 97.8     & 83.1      & 96.6     & 98.3     & 47.0     & 78.0     & 87.0     & 61.6     & 86.8     & 92.6     \\
				\textbf{TGDT-L }   & 78.5     & 95.8     & 98.2     & 90.1      & 98.6     & 99.8     & 58.3     & 82.9     & 90.1     & 74.8     & 92.5     & 95.9     \\
				\textbf{TGDT-GL }    & \textbf{79.1}     & \textbf{96.4}     & \textbf{98.7}     & \textbf{90.9}      & \textbf{99.1}     & \textbf{99.8}     & \textbf{58.9}     & \textbf{83.5}     & \textbf{90.4}     & 75.3     & \textbf{93.0}     & \textbf{96.3}     \\ \bottomrule
			\end{tabular}
		}
	\vspace{-1 em}
	\end{center}
	\label{tab:COCOoscar}
\end{table*}
\begin{figure*}
	\centering
	\subfigure[R@1 of TGDT-G]{
		\label{fig:param_analysis:a}
		\includegraphics[width=1.58in]{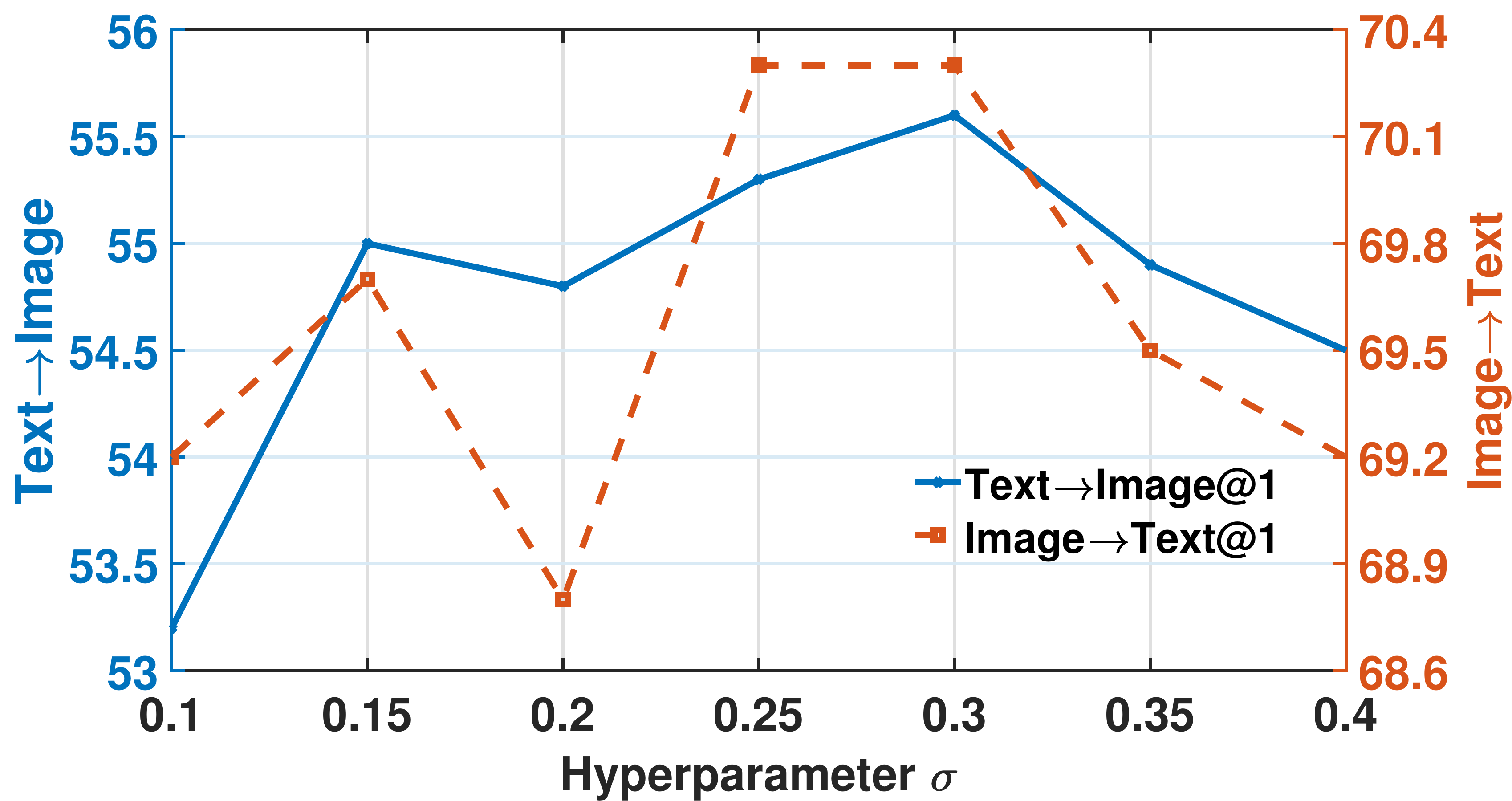}}
	\hspace{0.00in}
	\subfigure[R@10 of TGDT-G]{
		\label{fig:param_analysis:b}
		\includegraphics[width=1.58in]{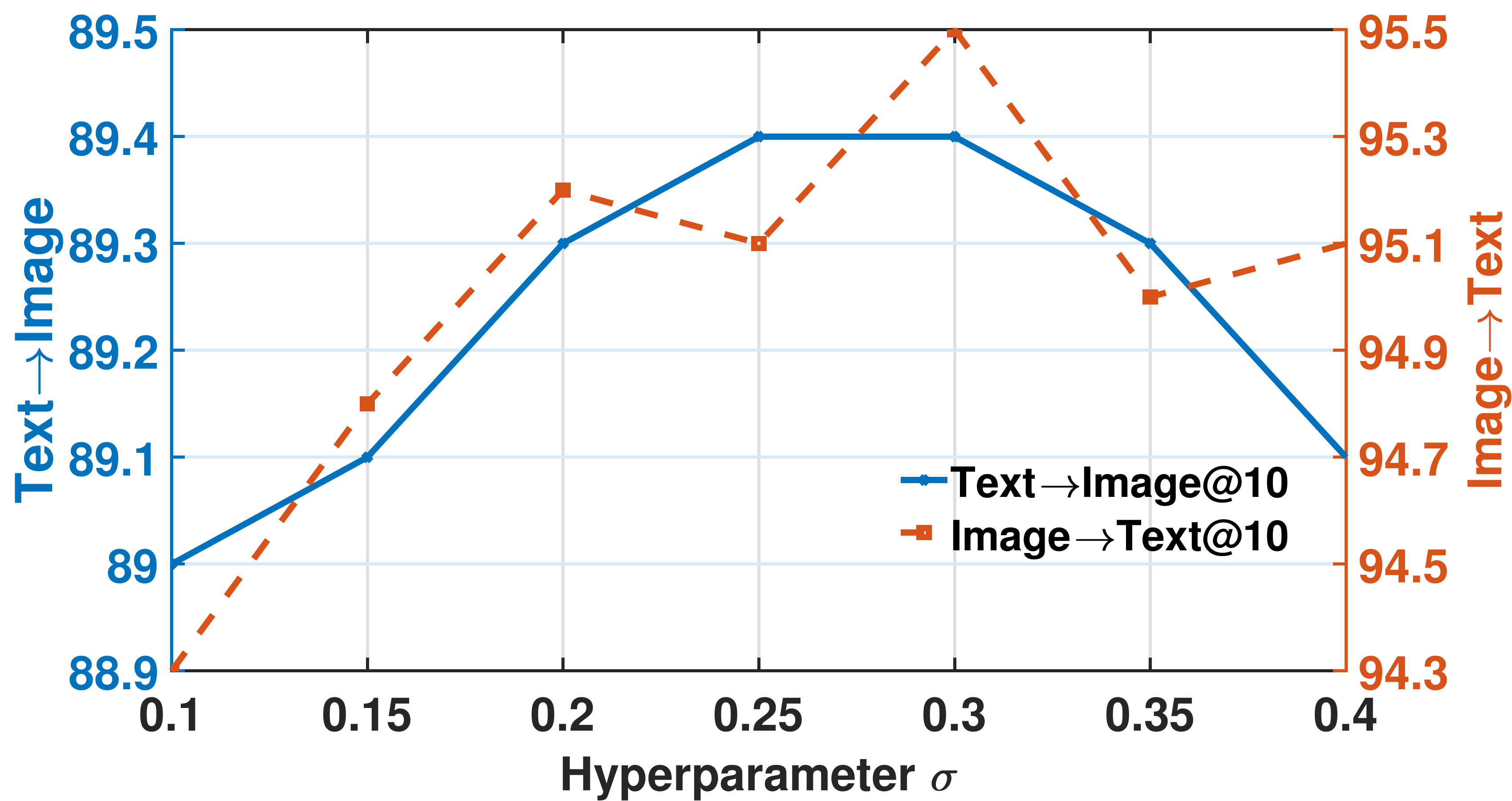}}
	\hspace{0.00in}
	\subfigure[R@1 of TGDT-L]{
		\label{fig:param_analysis:c}
		\includegraphics[width=1.58in]{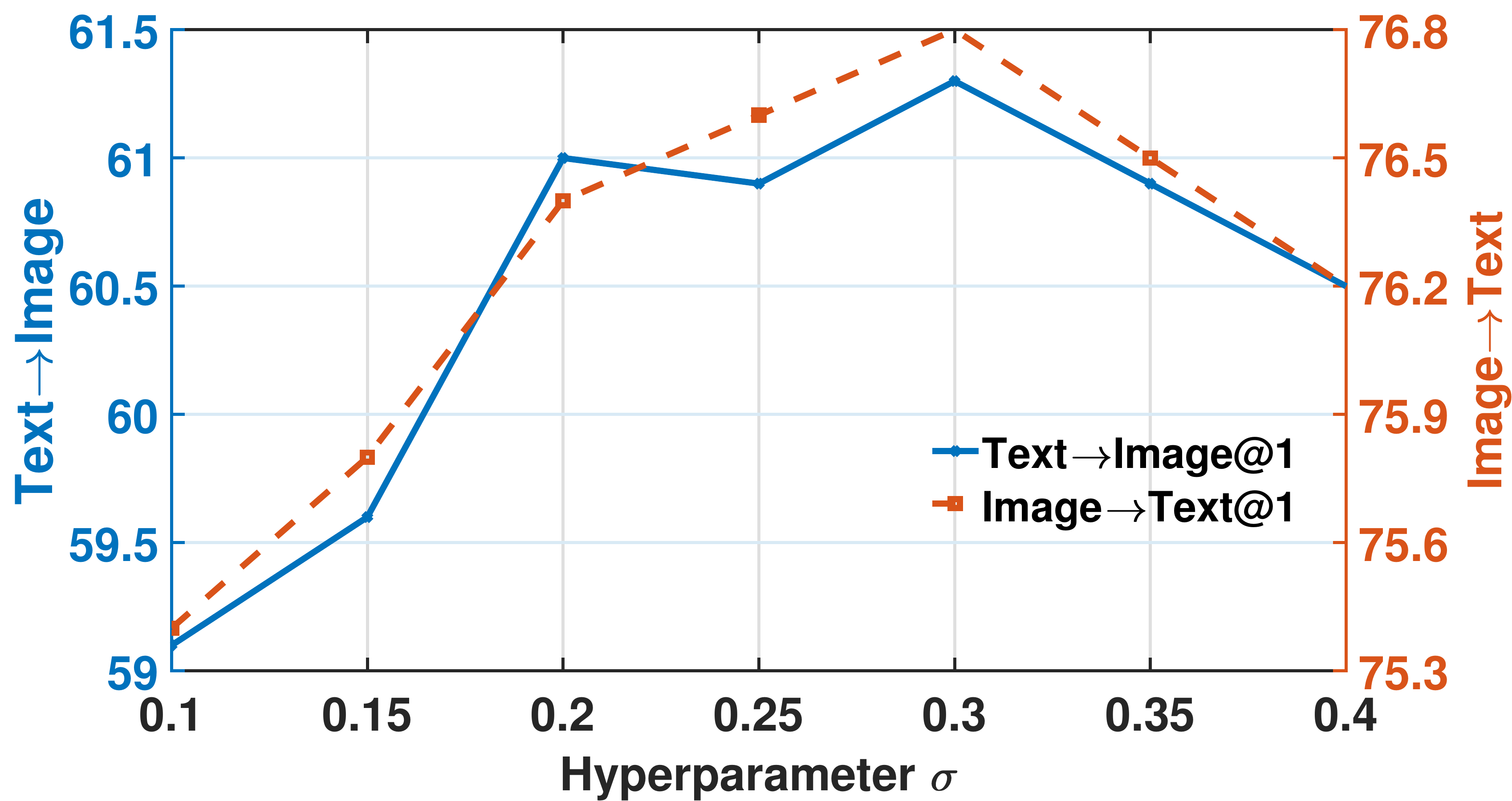}}
	\hspace{0.00in}
	\subfigure[R@10 of TGDT-L]{
		\label{fig:param_analysis:d}
		\includegraphics[width=1.58in]{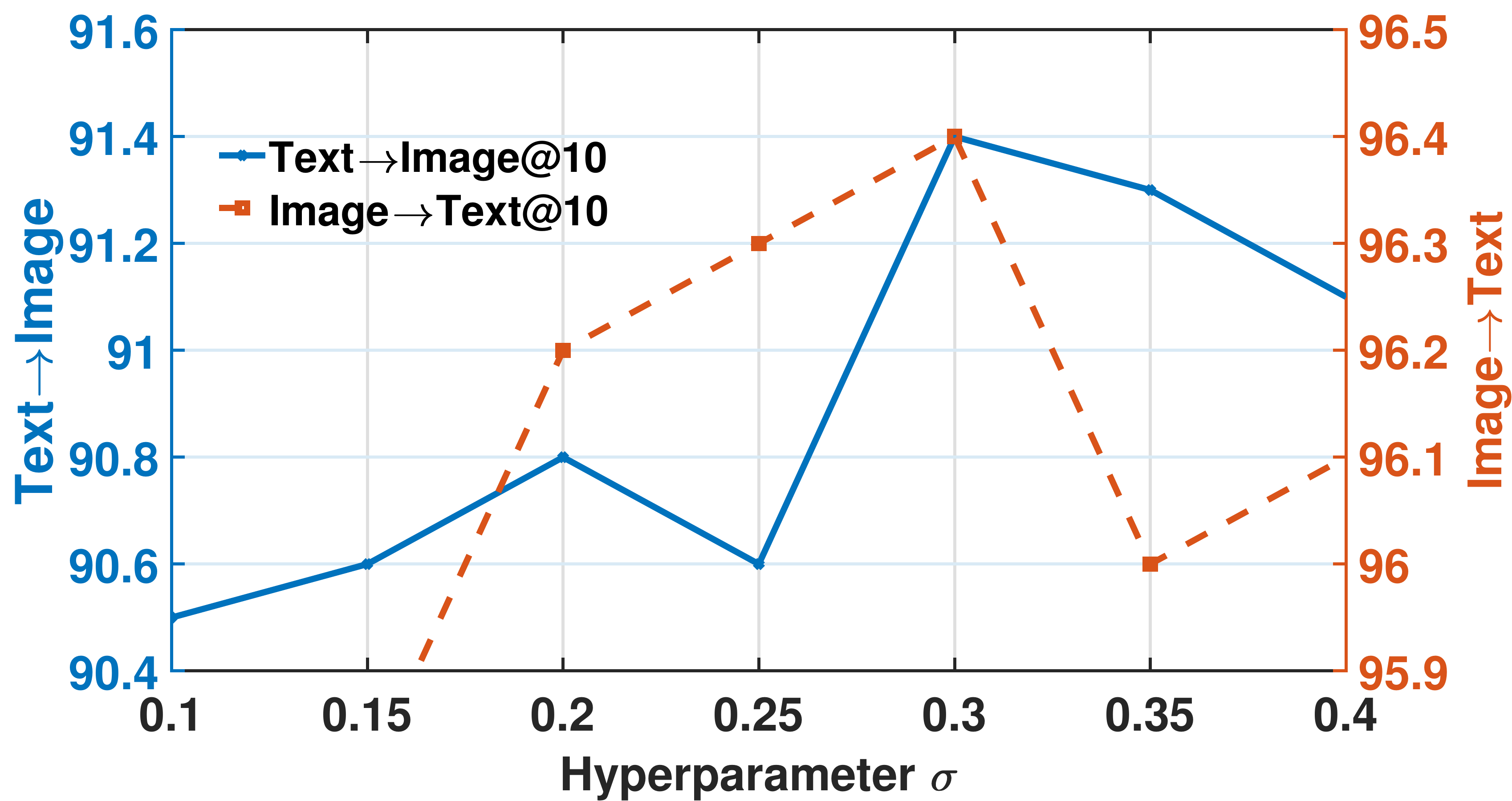}}
	\caption{The impact of $\sigma$ in the proposed CMC loss on retrieval performances on the Flickr30K dataset. This parameter is analyzed for both coarse- and fine-grained retrievals which use global and local representations, respectively. }
	\label{fig:param_sigma}
\vspace{-1 em}
\end{figure*}

\begin{figure}[t]
	\begin{center}
		\includegraphics[width=0.92\linewidth]{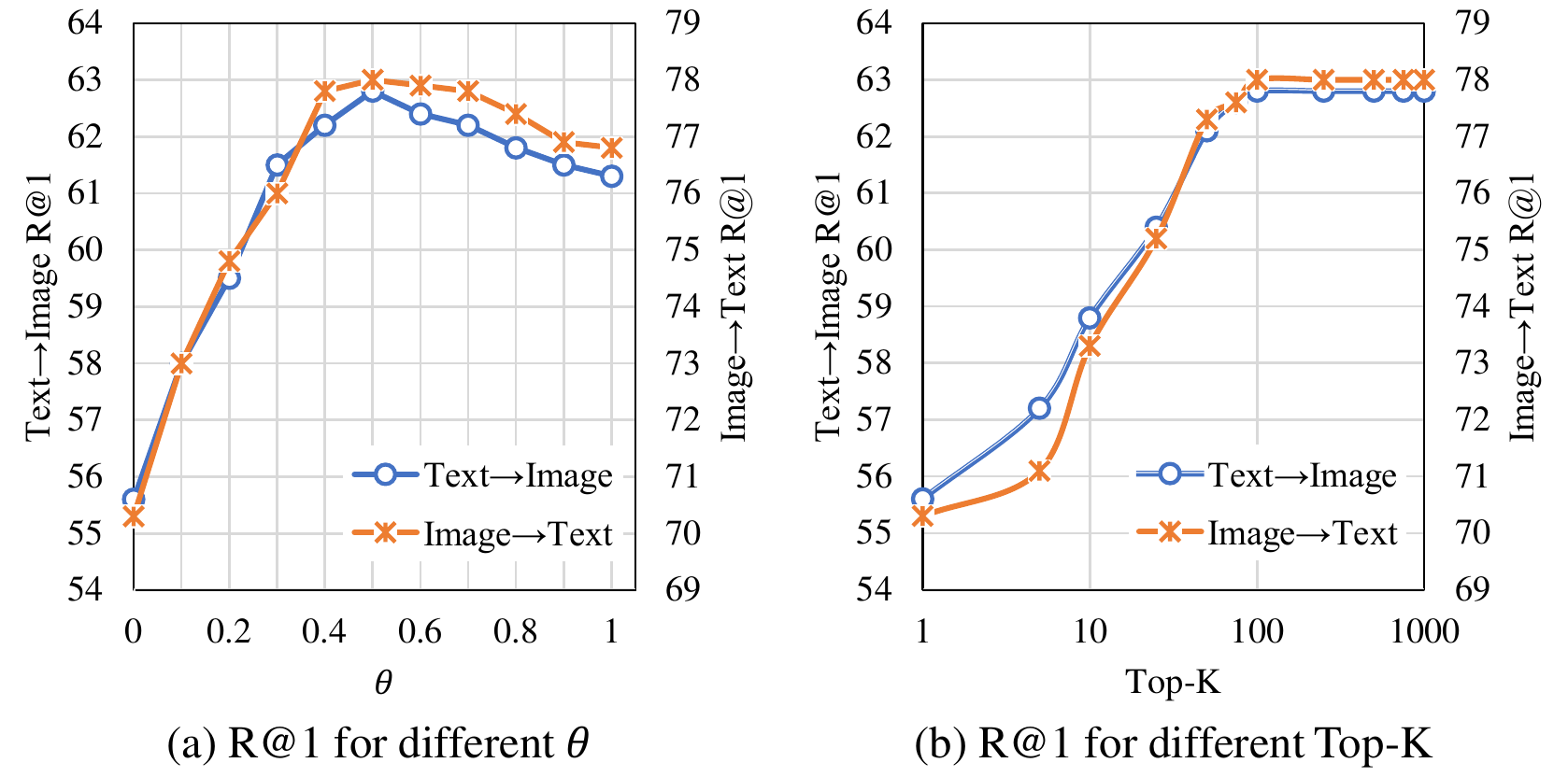}
	\end{center}
\vspace{-1 em}
	\caption{Retrieval results under different $\theta$ and Top-K on the Flickr30K dataset.} 
	\label{fig:abf}
\vspace{-0.5 em}
\end{figure}

\subsection{Comparison with VLP Models}
In recent years, Vision-Language Pre-training (VLP) models~\cite{CLIP,ALIGN,UnicoderVL,Oscar,VinVL} have been used in image-text retrieval tasks. These VLP based methods show very competitive performance against the coarse-grained or fine-grained retrieval methods (i.e., methods without VLP). One disadvantage of these methods is that it requires a huge amount of training data, which is laboriously
expensive and computationally unfeasible in most applications.

As an alternative, we use the VLP models to extract semantic features before training the TGDT architecture. More specifically, the pre-trained VinVL~\cite{VinVL} is used to extract image and text features. It should be emphasized that the VinVL encoder employs cross-attention operation and requires image-text pairs as the input. In contrast, our approach comprises two independent branches which receive image and text features, respectively. For simplicity, \textless Image,Text\textgreater denotes an image-text pair. We use \textless Mask,Image\textgreater\ and \textless Text,Mask\textgreater\ as the input of the VinVL encoder to extract image and text features, respectively, where $\mathrm{Mask}$ operation masks off all the elements for this modality. 
In this way, we can obtain the corresponding VLP features $V$ and $L$ for the image and text modalities, respectively. Through dual transformer encoders $\mathrm{ITR}$ and $\mathrm{TTR}$, global and local representations for each modality are subsequently learned. The dual transformer encoders are trained on different datasets with the proposed CMC loss function. Similar to Subsection~\ref{subsection:results_1}, we provide results of TGDT-G, TGDT-L, and TGDT-GL, respectively. 

Tab.~\ref{tab:F30Koscar} shows results on the Flickr30K dataset. For both image-to-text retrieval, and text-to-image retrieval, the proposed TGDT-GL considerably outperforms recent VLP models. 

The results on the COCO 1K and COCO 5K are summarized in Tab.~\ref{tab:COCOoscar}. Similarly, TGDT-GL yields the state-of-the-art performance for most evaluations, demonstrating obvious superior performances over the recent counterpart methods such as ViSTA-B~\cite{cheng2022vista}, ViLT-B~\cite{kim2021vilt} and SAVPVLP~\cite{xue2021probing}.


Compared with VinVL~\cite{VinVL} that adopts cross-attention encoders, our approach uses two independent image and text encoders and demonstrates superior performance for image-text retrieval. Notably, the inference speed of our method is much faster than that of VinVL, which would be experimentally proved in the following analyses. These experiments show that our approach can be combined with existing VLP methods to achieve even better performances. 

\subsection{Ablations and Analyses}
We further provide detailed experimental analyses, including sensitivity analysis of hyperparameters, inference speed of the cross-modal retrieval and ablation studies. 

\begin{figure}[t]
	\begin{center}
		\includegraphics[width=0.8\linewidth]{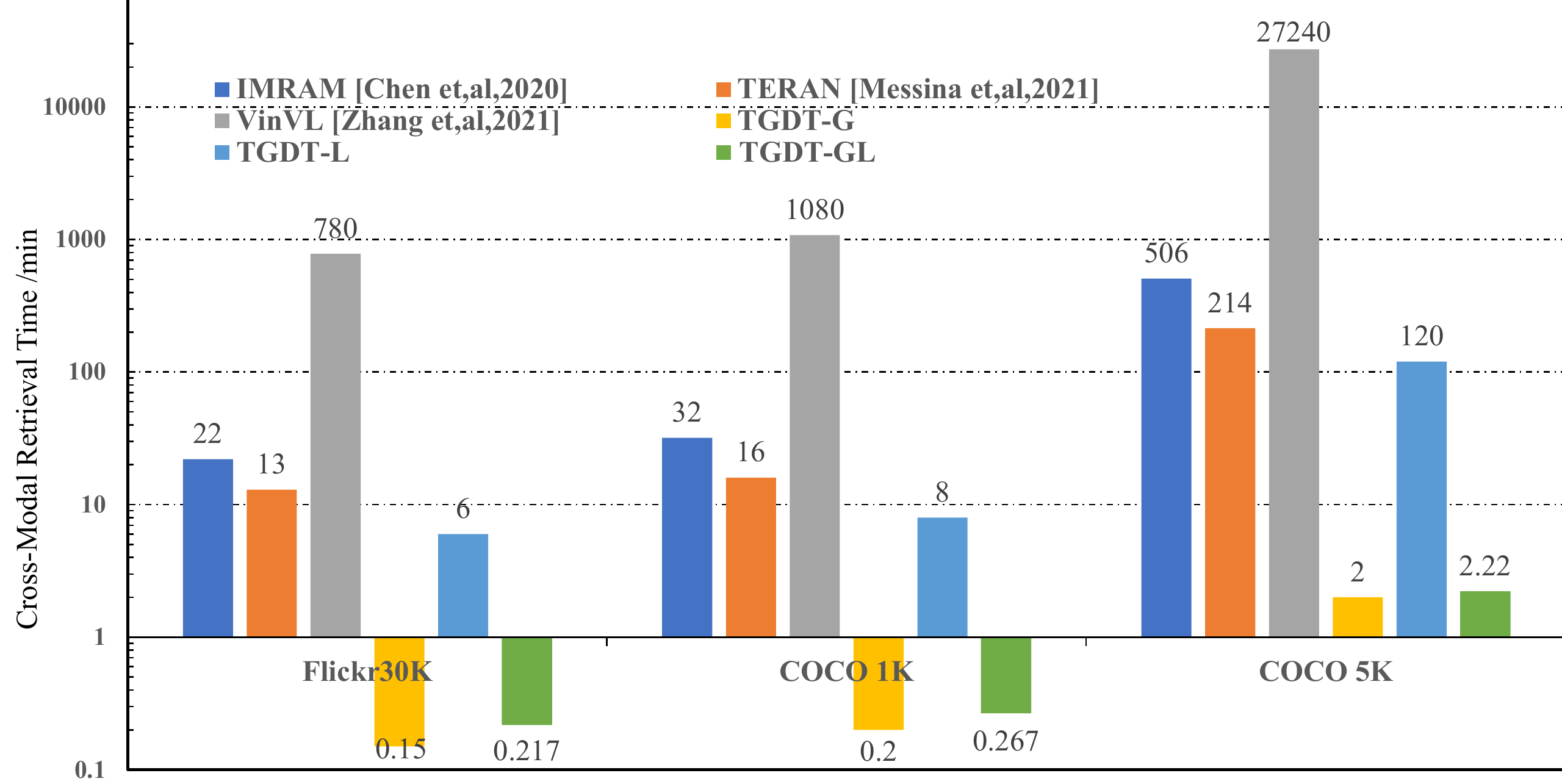}
	\end{center}
\vspace{-0.5 em}
	\caption{Inference time of representative methods on different datasets. The models run on an NVIDIA Tesla P100 GPU. Our method has a considerable speed advantage compared with current state-of-the-art methods.} 
	\label{fig:time}
\vspace{-0.5 em}
\end{figure}

\begin{table}[t]
	\caption{Ablation analysis of our proposed method on the Flickr30K dataset. The results verify the effectiveness of the two modules: JGLR and CMC.}
	\vspace{-2 em}
		\begin{center}
			\resizebox{1\linewidth}{!}{
				\begin{tabular}{lccccccccc}
					\toprule
					\multirow{2}{*}{Method} & \multirow{2}{*}{G} &  \multirow{2}{*}{L}
					& \multicolumn{3}{c}{Text→Image}                & \multicolumn{3}{c}{Image→Text}                \\
					& &  & R@1           & R@5           & R@10          & R@1           & R@5           & R@10         
					\\ \midrule
					Baseline-G & \checkmark &  & 48.2          & 76.7          & 84.6          & 61.3          & 85.1          & 90.6          \\
					TGDT-G w/o JGLR  & \checkmark &  & 51.6          & 80.0          & 87.1          & 66.4          & 86.7          & 92.0          \\
					TGDT-G w/o CMC & \checkmark &  & 53.1          & 81.6          & 88.7          & 67.9          & 87.0          & 92.3          \\
					TGDT-G & \checkmark &  & \textbf{55.6} & \textbf{83.1} & \textbf{89.4} & \textbf{70.3} & \textbf{91.4} & \textbf{95.5} \\ 
					\midrule
					Baseline-L &  & \checkmark & 54.4          & 81.5          & 88.2          & 68.1          & 88.6          & 93.7          \\
					TGDT-L w/o JGLR &  &  \checkmark& 56.9          & 83.8          & 89.4          & 72.1          & 89.8          & 94.0          \\
					TGDT-L w/o CMC  &  & \checkmark & 58.8          & 84.8          & 91.0          & 74.4          & 91.6          & 95.7          \\
					TGDT-L &  & \checkmark & \textbf{61.3} & \textbf{86.0} & \textbf{91.4} & \textbf{76.8} & \textbf{93.2} & \textbf{96.4} \\ 
					\bottomrule
				\end{tabular}
			}
		\end{center}
	\vspace{-1 em}
		\label{tab:ab}
	\end{table}
	
	\noindent \textbf{Hyperparameter analysis\ } 
	We analyze the impact of important hyperparameters of the proposed approach on retrieval performance. The main hyperparameters are the slack parameter $\sigma$ in the proposed CMC loss function, the ratio parameter $\theta$ in the two-stage inference, and the number of $K$ for the top-K candidate samples selection. It should be noted that the margin hyperparameter $\delta$ in the previous multimodal contrastive loss is not evaluated as it is usually set to 0.2 by practice.
	
	Fig.\ref{fig:param_sigma} shows the influence of $\sigma$ for both coarse- and fine-grained retrievals. The value of $\sigma$ is chosen from ${\rm{\{ 0}}{\rm{.1,0}}{\rm{.15}} \cdots {\rm{0}}{\rm{.4\} }}$. We can see the best performance is reached when $\sigma = 0.3$ for both text-to-image retrieval and image-to-text retrieval. When $\sigma < 0.3$, retrieval accuracy monotonically increases with the increase of $\sigma$.
	
	The impact of $\theta$ in the two-stage inference is presented in Fig.~\ref{fig:abf} (a). It can be seen that $\theta=0.5$ yields the best performance. The best performance is dramatically higher than those when $\theta=0$ or $\theta=1$, which denotes the inference strategies that only use global similarity or local similarity. 
	\RED{It should be noted that two-stage inference based on fused features beats single-stage inference based on local features for both efficiency and accuracy. The reason for efficiency improvement comes from the using of two-stage inference, while the reason for accuracy improvement is that fused features are better than either local features or global features. The results indicate that global features that describe the overall semantic information of the entire image or sentence and local features which represent candidate object regions or segments of the sentence are both effective and complementary with each other for image-text retrieval.}
	
	Fig.~\ref{fig:abf} (b) tests the effect of $K$ on retrieval performance after re-ranking. The best performance is reached when K is close to 100. The performance monotonically increases when $K$ increases from 1 to 100, and gets saturated when $K > 100$.
	\vspace{0.2 em}
	
	\noindent \textbf{Comparison of Inference Time\ } 
	For the test subset of a particular dataset, we count the total inference time of the two tasks of image-to-text and text-to-image retrievals. The representative state-of-the-art methods are IMRAM~\cite{IMRAM}, TERAN~\cite{TERAN}, VinVL~\cite{VinVL}. The former two are fine-grained retrieval methods, and the latter one is VLP method. The early coarse-grained retrieval methods are not selected due to low retrieval performances.
	
	The results on different datasets are summarized in Fig.~\ref{fig:time}. Our methods have a considerable speed advantage over the current best-performing methods. Consistent with our expectations, TGDT-G requires the shortest time to complete the retrieval task. TGDT-GL selects the top 100 global similarity samples for re-ranking and obtains a considerable performance improvement under the premise of a slight loss of speed.
	\vspace{0.2 em}
	
	\noindent \textbf{Ablation analysis\ } 
	The proposed TGDT architecture mainly consists of two novel components: Joint Global and Local Retrievals (JGL) and Consistent Multimodal Contrastive (CMC) loss. We ablated each of these components to analyze their effect on the retrieval performance. The baseline is the method that uses the same network structure as TGDT, but global and local retrieval tasks are trained separately, and the previous multimodal contrastive loss is used during training. The abbreviations w/o JGLR and  w/o CMC denote that separately training global and local retrieval tasks and using the previous multimodal contrastive loss, respectively. Without loss of generality, we only give the results on the Flickr30K dataset in Tab.~\ref{tab:ab}. 
	
	Taking the global retrieval as an example, TGDT-G w/o JGLR improves the Baseline-G for Text→Image@1 and Image→Text@1 by 3.4\% and 5.1\%, respectively. This demonstrates the effectiveness of the proposed CMC loss for cross-modal training.
	\RED{It is can be explained that the CMC loss which combines both intra-modal and inter-modal contrastive learning ensures semantic distance consistency between unmatched samples, and this consistency helps the model learn more discriminative features.}
	In addition, TGDT-G also improves TGDT-G w/o JGLR for Text→Image@1 and Image→Text@1 by 4.0\% and 3.9\%, respectively, which also confirms the benefits of the JGL module. 
	\RED{The results indicate that joint learning of global and local features improves the semantic representation ability of both types of features, thereby improving the retrieval accuracy.}
	
\section{Conclusion}
This paper presents a Token-Guided Dual Transformer (TGDT) architecture and a Consistent Multimodal Contrastive (CMC) loss for image-text retrieval. The TGDT promotes learning fast image-text retrieval models with the guidance of token-level cross-modal alignment. The CMC loss ensures the consistency of both local and global similarities between multimodal samples. Comprehensive experiments show that TGDT outperforms the state-of-the-art approaches for both text-to-image retrieval and image-to-text retrieval. Further experiments reveal that equipped with features extracted by the Vision-Language Pre-training (VLP) models, the proposed TGDT beats the state-of-the-art of VLP models for image-text retrieval. In addition, the TGDT gains remarkable advantages over most fine-grained retrieval approaches as well as the VLP methods in terms of inference speed. 
\RED{One limitation of the proposed method is that it uses the pre-trained Faster R-CNN to extract to local visual features, which may hinder the network to learn rich semantic features. To address this limitation, we will jointly train the backbone and the proposed TGDT for feature extraction and cross-modal feature alignment.}

%
\IEEEpeerreviewmaketitle

\ifCLASSOPTIONcaptionsoff
  \newpage
\fi

\bibliographystyle{IEEEtran}
\bibliography{egbib}

\end{document}